\documentclass[runningheads]{llncs}
\usepackage[T1]{fontenc}
\usepackage{graphicx}
\usepackage{mathtools}
\usepackage{amssymb}
\usepackage{booktabs}
\usepackage[linesnumbered,ruled,vlined]{algorithm2e}
\SetAlFnt{\scriptsize}  
\SetAlCapFnt{\scriptsize}
\SetAlCapNameFnt{\scriptsize}

\usepackage{float}
\usepackage{flafter}
\usepackage{afterpage}

\usepackage[framemethod=TikZ]{mdframed}

\usepackage{placeins}
\usepackage{multirow}

\usepackage{tikz}
\usetikzlibrary{positioning, shapes, arrows, petri}

\usepackage{subfiles}
\usepackage{subfig}

\usepackage{hyperref}
\extrafloats{100}
\SetKwComment{Comment}{/* }{ */}
\usepackage{lineno}
\usepackage{xcolor}

\newcommand{\preset}[1]{{}^{\bullet}{#1}}
\newcommand{\postset}[1]{{#1}^{\bullet}}

\newcommand{\incite}[2]{#1 #2}

\newcommand{\term}[1]{\emph{#1}}
\newcommand{\mi}[1]{\mathit{#1}}

\newcommand{\ac}[1]{\texttt{\small\text{#1}}}

\usepackage[normalem]{ulem}

\newcommand{\ch}[1]{#1}
\newcommand{\ad}[1]{#1}

\newcommand{\toch}[1]{#1}
\usepackage{todonotes}\setuptodonotes{inline}

\newcommand{\astar}{Astar\xspace}
\newcommand{\dijk}{Dijkstra\xspace}

\usepackage{subcaption} 

\newcommand{\mysubsubsection}[1]{\vspace{0.5em}\noindent\textbf{#1}}

\usepackage{microtype}

\newcommand{\acmakebook}{\ac{MakeBk}\xspace}
\newcommand{\acsubmitpay}{\ac{SubmitPD}\xspace}
\newcommand{\acawait}{\ac{AwaitC}\xspace}
\newcommand{\accountersign}{\ac{Sign}\xspace}
\newcommand{\acsubmitenroll}{\ac{SubmitPoE}\xspace}

\newcommand{\folda}{FoldA}
\newcommand{\foldan}{\folda$_n$\xspace}
\newcommand{\foldah}{\folda$_h$\xspace}

\begin{document}
\title{FoldA: Computing Partial-Order Alignments \\Using Directed Net Unfoldings}
%
%
\author{Douwe Geurtjens\inst{1}
\and Xixi Lu\inst{1}\orcidID{0000-0002-9844-3330}}
%
\authorrunning{Geurtjens and Lu}
%
\institute{Utrecht University, Utrecht, the Netherlands \\
\email{\{d.geurtjens, x.lu\}@uu.nl}}
%
\maketitle              
\begin{abstract}
Conformance checking is a fundamental task of process mining, which quantifies the extent to which the observed process executions match a normative process model.
The state-of-the-art approaches compute \textit{alignments} by exploring the state space formed by the synchronous product of the process model and the trace. This often leads to state space explosion, particularly when the model exhibits a high degree of choice and concurrency. Moreover, as alignments inherently impose a sequential structure, they fail to fully represent the concurrent behavior present in many real-world processes. To address these limitations, this paper proposes a new technique for computing partial-order alignments \emph{on the fly} using directed Petri net unfoldings, named \folda. We evaluate our technique on 485 synthetic model-log pairs and compare it against \astar- and \dijk-alignments on 13 real-life model-log pairs and 6 benchmark pairs. The results show that our unfolding alignment, although it requires more computation time, generally reduces the number of queued states and provides a more accurate representation of concurrency.


\end{abstract}

\section{Introduction}\label{sec:introduction}


Conformance checking assesses how well real-world executions adhere to a predefined process model~\cite{van2022process}. 
Various methods have been developed for this task, with one of the most widely used being \emph{alignment}-based techniques. These techniques compute an optimal sequence of model-allowed activities that minimizes deviations from a given trace~\cite{adriansyah2014aligning}. 
The alignment-based techniques have been extended in many different ways, such as data-aware~\cite{DBLP:journals/computing/MannhardtLRA16} and object-centric methods~\cite{DBLP:conf/caise/GianolaMW24}.


Current alignment computation relies on exploring the reachability graph of the synchronous product between an input trace and a predefined process model~\cite{adriansyah2014aligning}.  
However, when the model exhibits a high degree of concurrency, the corresponding reachability graph often suffers from state-space explosion due to interleaving semantics, causing it to grow to massively~\cite{mcmillan1993using}. While alignment techniques use graph search algorithms (e.g., \dijk and \astar) and heuristics to reduce the number of explored paths~\cite{adriansyah2014aligning}, they still select an arbitrary optimal sequential path in the reachability graph and return this as optimal alignment. 

%

Furthermore, these techniques impose a sequential order on the events of a trace and, consequently, on the returned alignment, even when no explicit dependencies exist. This limitation prevents them from accurately capturing the concurrent behavior present in many real-world processes~\cite{lu2015conformance}. \emph{Partial-order alignments} were introduced to address this issue and to identify discrepancies in
dependencies~\cite{lu2015conformance}. \ad{The partial-order alignments naturally capture concurrency and dependencies between moves, enabling the identification of much richer deviating patterns while preserving explicit relations between the moves~\cite{DBLP:journals/kais/LeemansZL23,DBLP:journals/tse/Garcia-Banuelos18}. For example,~\cite{DBLP:conf/etfa/LuMFA14} shows their diagnostic value in a healthcare setting. 

While partial-order alignments can be used to derive better diagnostics~\cite{DBLP:conf/etfa/LuMFA14,DBLP:journals/tse/Garcia-Banuelos18}, their computation \emph{still} relies on} sequential exploration of the reachability graph to compute an optimal alignment, which can lead to suboptimal results when dependencies are considered.

\begin{figure}[tb]
    \centering
    \begin{minipage}{0.45\textwidth} 
        \centering
        \vspace{0pt} 
        \resizebox{1.1\textwidth}{!}{
            \begin{tikzpicture}[
    place/.style={circle, draw=blue, thick, minimum size=3mm},
    transition/.style={rectangle, draw=black, thick, minimum size=4mm},
    sync/.style={rectangle, draw=blue, fill=black, thick, minimum width=3mm, minimum height=8mm},
    node distance=0.5cm and 0.6cm, 
    >=latex 
]

\node (p1) [place] {};
\node (sync1) [transition, right=of p1, label=center:S] {};

\node (p2a) [place, above right=0.1cm and 0.4cm of sync1] {}; 
\node (p2b) [place, below right=0.1cm and 0.4cm of sync1] {};

\node (A1) [transition, right=0.35cm of p2a, label=center:A] {};
\node (B1) [transition, right=0.35cm of p2b, label=center:B] {};

\node (p3a) [place, right=0.35cm of A1] {}; 
\node (p3b) [place, right=0.35cm of B1] {};

\node (sync2) [transition, below right=0.1cm and 0.4cm of p3a, label=center:C] {};
\node (p4) [place, right=of sync2] {};

\node (A2) [transition, below=of p2b, label=center:A] {};
\node (pS) [place, left=0.35cm of A2] {};
\node (S2) [transition, left=0.35cm of pS, label=center:S] {};
\node (p5) [place, right=0.35cm of A2] {};
\node (B2) [transition, below=of p3b, label=center:B] {};
\node (Bp) [place, right=0.35cm of B2] {};
\node (C2) [transition, right=0.35cm of Bp, label=center:C] {};

\node[above=0.001cm of sync1] {$t_1$};
\node[above=0.001cm of A1] {$t_2$};
\node[above=0.001cm of B1] {$t_3$};
\node[above=0.001cm of sync2] {$t_4$};

\node[above right=0.001cm and 0.001cm of S2] {$t_5$};
\node[above=0.001cm of A2] {$t_6$};
\node[above=0.001cm of B2] {$t_7$};
\node[above left=0.001cm and 0.001cm of C2] {$t_8$};

\draw[->, thick] (p1) -- (sync1);
\draw[->, thick] (sync1) -- (p2a);
\draw[->, thick] (sync1) -- (p2b);
\draw[->, thick] (p2a) -- (A1);
\draw[->, thick] (p2b) -- (B1);
\draw[->, thick] (A1) -- (p3a);
\draw[->, thick] (B1) -- (p3b);
\draw[->, thick] (p3a) -- (sync2);
\draw[->, thick] (p3b) -- (sync2);
\draw[->, thick] (sync2) -- (p4);

\draw[->, thick] (p1) -- (S2);
\draw[->, thick] (S2) -- (pS);
\draw[->, thick] (pS) -- (A2);
\draw[->, thick] (A2) -- (p5);
\draw[->, thick] (p5) -- (B2);
\draw[->, thick] (B2) -- (Bp);
\draw[->, thick] (Bp) -- (C2);
\draw[->, thick] (C2) -- (p4);

\end{tikzpicture}
            \hspace{0.5cm}\begin{tikzpicture}[
    place/.style={isosceles triangle, draw=black, thick, minimum size=5mm},
    node distance=0.5cm and 0.6cm, 
    >=latex 
]

\node (S1) [place, label=center:S] {};
\node (A1) [place, above right=0.2cm and 0.4cm of S1, label=center:A] {};
\node (B1) [place, below right=0.2cm and 0.4cm of S1, label=center:B] {};
\node (sync2) [place, below right=0.2cm and 0.6cm of A1, label=center:C] {};

\node[above=0.001cm of S1] {$e_1$};
\node[above=0.001cm of A1] {$e_2$};
\node[above=0.001cm of B1] {$e_3$};
\node[above=0.001cm of sync2] {$e_4$};

\draw[->, thick] (S1) -- (A1);
\draw[->, thick] (S1) -- (B1);
\draw[->, thick] (A1) -- (sync2);
\draw[->, thick] (B1) -- (sync2);

\end{tikzpicture}
        }
        \vspace*{\fill}
        \captionof{figure}{A Petri net $M$ (left) and \\a partial-order trace $\sigma$ (right).}
        \label{fig:example_net}
    \end{minipage}
    \hfill
    \begin{minipage}{0.53\textwidth} 
        \centering
        \resizebox{0.7\textwidth}{!}{


\begin{tikzpicture}[
    place/.style={circle, draw=black, thick, minimum size=3mm},
    modelplace/.style={circle, draw=blue, thick, minimum size=3mm},
    transition/.style={rectangle, draw=black, thick, minimum size=4mm},
    sync/.style={rectangle, draw=black, fill=black, thick, minimum width=3mm, minimum height=8mm},
    node distance=0.5cm and 0.6cm, 
    >=latex 
]

\node (l1) [place] {};
\node (p1) [modelplace, below=of l1] {};

\node (s) [transition, below right=0.1cm and 0.4cm of l1] {S,S};

\node (pa) [modelplace, right=1.8cm of l1] {};
\node (la) [place, above=0.1cm of pa] {};
\node (lb) [place, right=1.8cm of p1] {};
\node (pb) [modelplace, below=0.1cm of lb] {};

\node (a) [transition, right=of pa] {A,A};
\node (b) [transition, right=of lb] {B,B};

\node (ap) [modelplace, right=of a] {};
\node (al) [place, above=0.1cm of ap] {};
\node (bl) [place, right=of b] {};
\node (bp) [modelplace, below=0.1cm of bl] {};

\node (c) [transition, right=4.2cm of s] {C,C};

\node (pc) [modelplace, right=2.2cm of ap] {};
\node (lc) [place, right=2.2cm of bl] {};

\node[above=0.001cm of s] {($t_1,e_1$)};
\node[above=0.001cm of a] {($t_2,e_2$)};
\node[below=0.001cm of b] {($t_3,e_3$)};
\node[below=0.001cm of c] {($t_4,e_4$)};

\node (s2) [transition, below=of s] {S,S};

\node (pa2) [modelplace, below=of pb] {};
\node (la2) [place, below=0.1cm of pa2] {};
\node (lb2) [place, below=0.1cm of la2] {};

\node (a2) [transition, right=of pa2] {A,A};

\node (al2) [place, right=of a2] {};
\node (ap2) [modelplace, below=0.1cm of al2] {};

\node (b2) [transition, right=2.8cm of lb2] {B,B};

\node (bl2) [place, right=of b2] {};
\node (bp2) [modelplace, above=0.1cm of bl2] {};



\node[below=0.001cm of s2] {($t_5,e_1$)};
\node[below=0.001cm of a2] {($t_6,e_2$)};
\node[above=0.001cm of b2] {($t_7,e_3$)};








\draw[->, thick] (l1) -- (s);
\draw[->, thick] (p1) -- (s);

\draw[->, thick] (s) -- (la);
\draw[->, thick] (s) -- (lb);
\draw[->, thick] (s) -- (pa);
\draw[->, thick] (s) -- (pb);
\draw[->, thick] (la) -- (a);
\draw[->, thick] (pa) -- (a);
\draw[->, thick] (lb) -- (b);
\draw[->, thick] (pb) -- (b);

\draw[->, thick] (a) -- (al);
\draw[->, thick] (a) -- (ap);
\draw[->, thick] (b) -- (bl);
\draw[->, thick] (b) -- (bp);

\draw[->, thick] (al) -- (c);
\draw[->, thick] (ap) -- (c);
\draw[->, thick] (bl) -- (c);
\draw[->, thick] (bp) -- (c);

\draw[->, thick] (c) -- (lc);
\draw[->, thick] (c) -- (pc);

\draw[->, thick] (l1) -- (s2);
\draw[->, thick] (p1) -- (s2);

\draw[->, thick] (s2) -- (pa2);
\draw[->, thick] (s2) -- (la2);
\draw[->, thick] (s2) -- (lb2);

\draw[->, thick] (pa2) -- (a2);
\draw[->, thick] (la2) -- (a2);

\draw[->, thick] (a2) -- (ap2);
\draw[->, thick] (a2) -- (al2);

\draw[->, thick] (ap2) -- (b2);
\draw[->, thick] (lb2) -- (b2);

\draw[->, thick] (b2) -- (bp2);
\draw[->, thick] (b2) -- (bl2);




\end{tikzpicture}

         }
        \vfill
        \captionof{figure}{An (incomplete) prefix unfolding of the synchronous product of $M$ and $\sigma$.}
        \label{fig:example_unfolding}
    \end{minipage}
\end{figure}

Consider the example model and an example of a partial-order trace (\ch{also known as a pomset}) in~\autoref{fig:example_net}. Current alignment techniques consider both 
$\gamma_1$ = {\scriptsize\begin{tabular}{|c|c|c|c|}
$e_1$ & $e_2$ & $e_3$ & $e_4$ \\ \hline
$t_1$ & $t_2$ & $t_3$ & $t_4$
\end{tabular}}
and $\gamma_2 = ${\scriptsize\begin{tabular}{|c|c|c|c|}
$e_1$ & $e_2$ & $e_3$ & $e_4$ \\ \hline
$t_5$ & $t_6$ & $t_7$ & $t_8$
\end{tabular}}
as optimal alignments and returns one arbitrarily. If $\gamma_2$ is chosen, the method in~\cite{lu2015conformance} uses it to compute a partial-order alignment and detects discrepancies in dependencies, for example, $A$ $(t_6)$ is directly followed by $B$ $(t_7)$ in the model, whereas in the trace, $A$ $(e_2)$ and $B$ $(e_3)$ are concurrent. However, the alternative alignment $\gamma_1$, which was not returned, would have produced a better partial-order alignment. 
%


In this paper, we propose a new approach, \emph{unfolding alignments}, leveraging the \emph{directed net unfoldings} technique. Net unfoldings naturally retain concurrency and dependencies in the original net. Pioneered by \incite{McMillan} \cite{mcmillan1993using}, the concepts of \ad{an adequate order and cut-off events} are introduced to mitigate the exponential growth of traditional state-space representations. Built on the idea of \ad{using an adequate order}, we introduce a cost-based, directed unfolding for computing alignments, allowing us to compute an optimal partial-order run (also called a \emph{configuration}) to the final marking in an \emph{on-the-fly} manner, see Fig.~\ref{fig:example_unfolding}. 

To evaluate the feasibility and performance of our unfolding alignments, we conducted experiments on 485 synthetic model-log pairs, 13 real-life model-log pairs, and 6 benchmark pairs. Due to the space constraints and the complexity of definitions, this paper focuses on sequential traces, allowing direct comparison with \dijk- and \astar-alignments. Nevertheless, \ch{the proposed unfolding alignment is also directly applicable to partial-order traces, see the example provided in Appendix~\ref{app:example}.} 
%
%
%
%
The remainder of the paper follows a standard structure. 
\section{Related Work}\label{sec:relatedwork}

\mysubsubsection{Conformance checking}
Early conformance checking research focused on defining quantitative metrics to evaluate process discovery algorithms \cite{greco2006discovering}. 
Replay-based conformance \cite{rozinat2008conformance} replays a trace on a model, artificially inserting tokens when transitions cannot fire, to identify the problematic places. However, this method struggled with invisible transitions and sometimes introduced behavior not allowed by the model, making it difficult to diagnose deviations \cite{van2012replaying,adriansyah2014aligning}. 
%
%
%
 To address these issues, the work in~\cite{van2012replaying} introduced alignments, which compute the optimal sequence of pairs of activities permitted by both the model and the trace while minimizing deviations. 
%
%
While alignments improved conformance checking accuracy, their sequential nature posed challenges for highly concurrent processes, as independent events could appear causally dependent \cite{lu2015conformance}. To address this, \incite{Lu et al.}{\cite{lu2015conformance}} proposed partial-order alignments, which allow non-sequential behavior by introducing partial-order traces. However, it still relies on the sequential alignment to return a sequential optimal alignment, which may not be the optimal partial-order alignment. 
%

\ad{García-Bañuelos et al~\cite{DBLP:journals/tse/Garcia-Banuelos18} \label{ln:relatedcompare} use net unfolding to obtain prime event structure (PES) for both the input model and the traces. After obtaining these PESs, they compare the two PESs by computing a partially synchronized product. To obtain PESs, they use the ERV unfolding~\cite{esparza1996improvement} to compute a complete prefix unfolding of the model, which means that it cannot handle unbounded, easy sound nets (i.e., a final marking is reachable). In contrast, our approach first builds the synchronous product of the model and trace, and then unfolds this synchronous product in a directed manner, allowing us to support unbounded, easy sound nets.}

\ad{Siddiqui et al.~\cite{mistake} also investigate the use of net unfolding for computing partial-order alignments. 
The idea in this paper was independently developed in 2024~\cite{douwethesis} and submitted before~\cite{mistake} became available. There are two key differences: (1) the work in~\cite{mistake} entirely omits the fundamental problem that current alignment-based approaches cannot compute optimal partial-order alignment due to unhandled discrepancies between moves. We highlight this problem explicitly with a concrete example in our problem statement. (2) Our evaluation is significantly broader. While \cite{mistake} tested only one model type (parallel branches, no loops) and a single real-life event log using only Inductive Miner, we evaluated 485 models with choice, parallelism, and loops, and 13 real-life log-model pairs with Inductive Miner and Split Miner, along with 6 benchmark sets. }

\mysubsubsection{Net Unfoldings}
Petri net unfoldings efficiently represent the state space of a net \cite{mcmillan1993using}. The unfolding of a Petri net, potentially infinite, is a Petri net without backward conflict. In such a net, \emph{no two transitions can output to the same place}, enforcing a strict partial order on transitions. In an unfolding, transitions are called \emph{events}, and places \emph{conditions}.

To ensure a compact representation, \incite{McMillan}{\cite{mcmillan1993using}} introduced \textit{cut-off events}, to stop the unfolding of an event when the marking is already seen in other branches of the unfolding.
Later, \incite{Esparza et al.,}{\cite{esparza1996improvement}} identified flaws in the original algorithm, where in some cases, it created an unfolding in which not all reachable markings were represented. They propose the concept of an \textit{adequate order} on the events of the unfolding, to guarantee that all possible markings are reached and improve computation efficiency. The use of cut-off events also leads to the concept of \textit{finite complete prefixes}, referring to the minimal part of the unfolding that represents all reachable markings. \

\incite{Bonet et al.,}{\cite{bonet2008directed}} refined the approach by generating only the prefix needed to reach a target marking, further enhancing performance.
Lastly, \incite{Bonet et al.,}{\cite{bonet2014recent}} provided an overview of the recent advances in Petri net unfoldings, covering both efficiency and applicability to different Petri net types.

\ad{Our approach builds the adequate order and cut-off event concepts from the ERV algorithm~\cite{mcmillan1993using}. However, unlike ERV, we do not compute a complete prefix unfolding. Instead, we perform a targeted, directed unfolding that prioritizes reaching the final marking as quickly as possible. This is achieved through a best-first search over the events to be appended, while respecting the adequate order. The unfolding process terminates as soon as the final marking is reached. }

\section{Preliminaries}\label{sec:preliminaries}
In this section, we formally define key terms relevant to our work. We assume our readers are familiar with \emph{multiset}, \emph{sequences}, and \emph{Petri nets}; otherwise, we provide definitions in Appendix~\ref{sec:notation} for review purposes. 

\vspace{0.5em}\noindent \textbf{Alignment based conformance checking.} 
We introduce the formal definitions related to event logs, process models, and alignments~\cite{adriansyah2014aligning}. 

\begin{figure}[b]
\centering
\includegraphics[width=\textwidth]{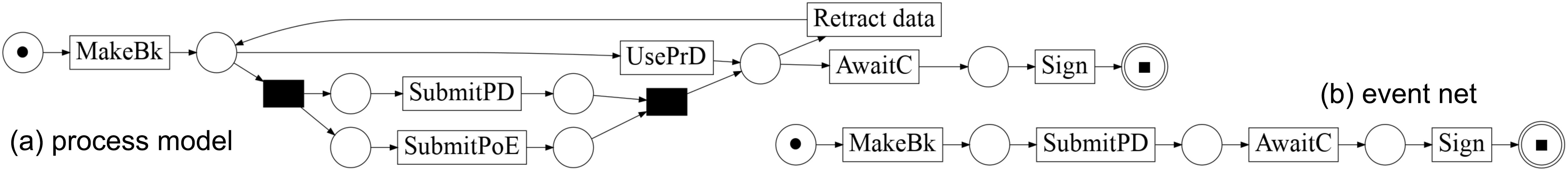}
\caption{\ch{(a) Process model $N_1$ in the form of a Petri net and (b) event net $N_l$ for the trace $\sigma_1$.}}
\label{fig:mod}
\end{figure}

Let $A$ be a set of activities. A \emph{trace} $\sigma = \langle e_1,e_2,...e_n \rangle \in A^*$ is a sequence of events where each event is the activity executed. An \emph{event log} $L \subset \mathcal{B}(A)$ is a multiset of traces. 
%
%
A process model $N_1$ in the form of a Petri Net for the student-housing application process is shown in Fig.~\ref{fig:mod}(a), and an example of trace is $\sigma_1 = \langle \acmakebook, \allowbreak \acsubmitpay, \allowbreak \acawait, \allowbreak \accountersign \rangle$ in an event net form in Fig.~\ref{fig:mod}(b).
%
Currently, only the \acmakebook transitions are enabled. 

\begin{definition}[Alignment] Let $N = (P, T, F, i, f)$ denote \ch{an \emph{easy sound Petri net} representing a process model} where $i$ is the initial marking and $f$ the final marking. Let $\sigma$ denote a trace. A valid alignment $\gamma \in ((A \cup \{\gg\}) \times (T \cup \{\gg\}))^*$ is a sequence of moves such that (1) $\pi_1(\gamma)_{\downarrow A} = \sigma$, \ch{i.e., projecting on the first elements of the moves yields the trace}, (2) $i \xrightarrow{\pi_2(\gamma)_{\downarrow T}} f$, \ch{i.e., projecting on the second elements yields an occurrence sequence from $i$ to $f$ in $N$}, and (3) $(\gg,\gg) \notin \gamma$.
For $a \in A$ and $t \in T$, we call $(a, \gg)$ a \term{log move}, $(\gg, t)$ a \term{model move}, $(a, t)$ a \term{synchronous move}, and $(\gg,\gg)$ an invalid move. 
\label{def:align}
\end{definition}
\vspace{-1em}

\begin{definition}[Optimal Alignment] Let $\Gamma_{\sigma, N}$ denote the set of all alignments between trace $\sigma$ and net $N$. Let $\kappa : (A\cup \{\gg\}) \times (T \cup \{\gg\}) \rightarrow \mathbb{R}^+$ be a \term{cost function} for moves. We say that $\gamma \in \Gamma_{\sigma, N}$ is an \term{optimal alignment} between $\sigma$ and $N$ if and only if for all $\gamma' \in \Gamma_{\sigma, N}, \kappa(\gamma) \leq \kappa(\gamma')$, where $\kappa(\gamma)= \sum_{(e,t)\in \gamma} \kappa(e,t)$.  
\end{definition}
\vspace{-0.5em}
\vspace{-0.5em}
\begin{definition}[Synchronous product] \ch{The \term{synchronous product} of two disjoint nets $N_l = (P_l, T_l, F_l, i_l, f_l)$ and $N_m = (P_m, T_m, F_m, i_m, f_m)$ is an easy sound \term{Petri net} $\mi{sp} = N_l \times N_m = (P, T, F, i, f)$, where
%
    (1) $P = P_l \cup P_m$; 
    (2) $T = (T_l\times\{ \gg\})  \cup (\{\gg\} \times T_m) \cup \mathit{SM}$, where $\mathit{SM} = \{(a, t) \in T_l \times T_m \mid a = t\}$;  
    (3)~$F = 
    \bigcup_{(t_l, t_m) \in T} 
    \{ (p_l, (t_l, t_m)) \mid (p_l, t_l) \in F_l \} 
    \cup \{ ((t_l, t_m), p_l) \mid (t_l, p_l) \in F_l \} 
    \cup \{ (p_m, (t_l, t_m)) \mid (p_m, t_m) \in F_m \} 
    \cup \{ ((t_l, t_m), p_m) \mid (t_m, p_m) \in F_m \}$,
    (4) $i = i_l \cup i_m$, and
    (5) $f = f_l \cup f_m$}. 
\end{definition}
\vspace{-0.5em}



\begin{figure}[b]
\centering
\makebox[\textwidth][c]{%
\includegraphics[width=1.2\textwidth]{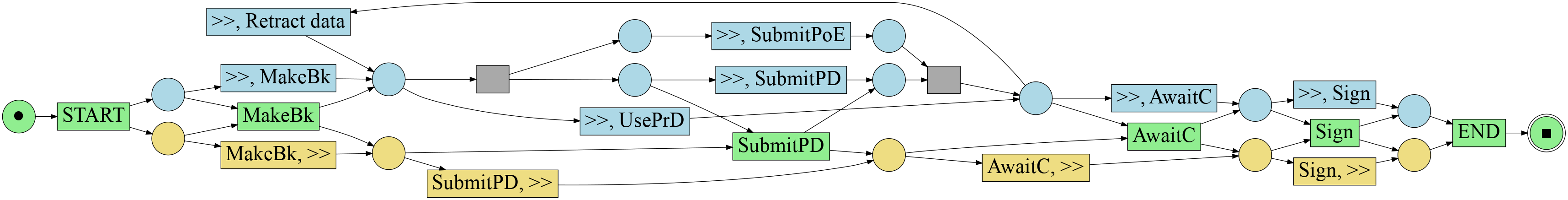}
}
\caption{\ch{The synchronous product between $N_1$ and $\sigma_1$ in Fig.~\ref{fig:mod}.}}
\label{fig:sp}
\end{figure}

Figure~\ref{fig:sp} shows the resulting synchronous product of $N_l$ and $N_1$, which is constructed as follows: the transitions present in the model become model moves (colored blue), the transitions in the event net become log moves (colored yellow), and for each pair of transitions that have the same label, we create a synchronous move transition  (colored green). 


By exploring the reachability graph of the synchronous product using a shortest path algorithm, we derive an optimal alignment $\gamma = \langle 
(\acmakebook, \allowbreak \acmakebook), \allowbreak (\gg, \acsubmitenroll),\allowbreak
(\acsubmitpay, \allowbreak 
\acsubmitpay),\allowbreak
(\acawait, \acawait),\allowbreak
(\accountersign, \accountersign) \rangle$.




\vspace{-0.5em}
\begin{definition}[Partial Order Alignment] Let $\mi{sp} = (P, T, F, i, f) = N_l \times N_m$ be a synchronous product of process model $N_m$ and event net $N_l$. A partial-order alignment $\delta$ is a partial-order over the moves such that (1), (2) and (3) from Definition \ref{def:align} \toch{are respected}.
\end{definition}
\vspace{-0.5em}

Informally, we can say that events in a partial-order trace can be ``tied" in their execution, rather than strictly sequential as in a conventional trace $\sigma$. The same holds for a partial-order alignment $\delta$ when compared to a regular alignment $\gamma$. In both cases, the only difference with their conventional counterpart is that the order is partial, rather than total.

\mysubsubsection{Net Unfoldings}
Here, we introduce the key definitions from~\cite{esparza1996improvement} to understand the process of unfolding a Petri Net, revisiting our running example. By ``unfolding'' a net $N$, we obtain a so-called branching process $\beta$ of $N$, which is a labeled occurrence net. \ad{Given a net, we say $x$ \emph{causes} $y$, i.e., $x < y$, if there is a directed path from $x$ to $y$; we say $x$ is \emph{in conflict with} $y$, i.e., $x\#y$, if $x$ and $y$ originate from the same place but follow diverging paths; and $x$ is \emph{concurrent} with $y$, i.e., $x || y$, if $x$ and $y$ are neither casually related nor in conflict.}


\vspace{-0.5em}
\begin{definition}[Occurrence net] An \term{occurrence net} is a net $O = (B, E, G)$ in which $B$ denotes a set of \term{conditions}, $E$ denotes a set of \term{events}, and $G$ denotes the causal relations such that:
(1) $\mid \preset{b} \mid \leq 1$ for every $b \in B$, meaning every condition has at most one incoming event;
(2) $O$ is acyclic; 
(3) $O$ is finitely preceded, \ad{i.e., for all $o \in O$, $o$ has a finite number of causal predecessors}; 
(4) No element is in conflict with itself.
\end{definition}


\vspace{-1em}
\begin{definition}[Branching process] A \term{branching process} $\beta$ of a net $N = (P, T, F, i, f)$ is a labeled occurrence net $(O, p) = (B, E, G, p)$ (possibly infinite) where the labeling function $p : (E \rightarrow T) \cup (B \rightarrow P)$ satisfies the following properties:
        (1) for every $e \in E$, the restriction of $p$ to $\preset{e}$ is a bijection between $\preset{e}$ in $N$ and $\preset{p}(e)$ in $\beta$, and similar for $\postset{e}$ and $\postset{p(e)}$;
        (2) $\beta$ starts with $i$, i.e., $p(\mi{Min}(O)) = i$ where $\mi{Min}(O) = \{b \in B | \preset{b} = \emptyset\}$;
        (3) for every $e_1, e_2\in E$, if $\preset{e_1} = \preset{e_2}$ then $e_1=e_2$ (i.e., $\beta$ does not duplicate events).
\end{definition}

\begin{figure}[tb]
\centering
\includegraphics[width=0.9\textwidth]{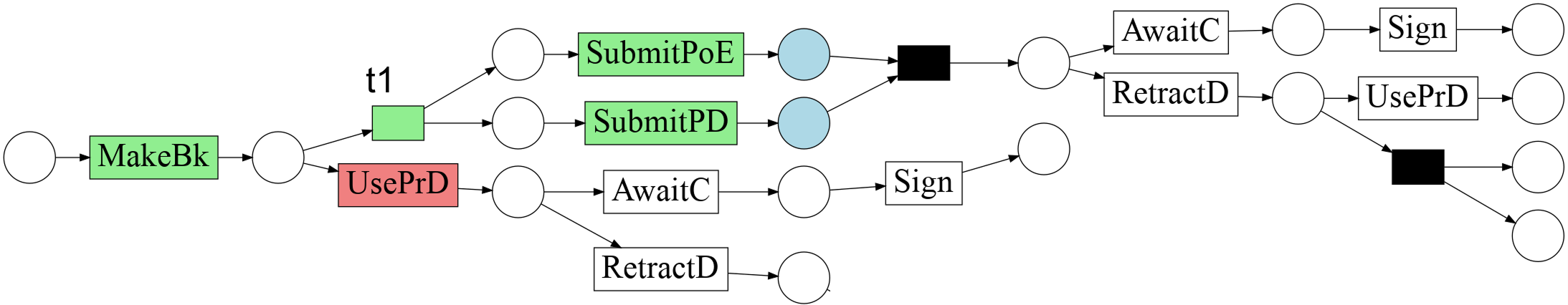}
\caption{A prefix of the unfolding of Figure~\ref{fig:mod}, with color coding of a configuration.}
\label{fig:unfconf}
\end{figure}

Figure~\ref{fig:unfconf} shows a branching process of the net in \autoref{fig:mod}. This unfolding has deliberately been cut off at an arbitrary point, as the full unfolding would be infinite. For the basic algorithm to construct an unfolding, we refer to \cite{mcmillan1993using}.


\vspace{-0.5em}
\begin{definition}[Configuration, Cut]
A \term{configuration} $C \subseteq E$ of a branching process $\beta = (O, p) = (B, E, G, p)$ is a set of events such that (1) $C$ is causally closed, \ad{i.e., $\forall e\in C, \forall e'<e : e' \in C$} and (2) $C$ is conflict-free, \ad{i.e., $\forall e, e' \in C : \neg(e\#e')$}. \ch{We use $\mi{Cut}(C)$ to denote the \term{cut} of a configuration, which is a set of conditions in $\beta$ that \emph{represents} the marking of the net after firing the events in the configurations, i.e., $\mi{Cut}(C) = ( \mi{Min}(O) \cup \postset{C}) \backslash \preset{C}$.} We use the \term{marking of configuration} $\mi{Mark}(C) = p(\mi{Cut}(C))$ to refer to the marking in the original net. We define $\uparrow\!\!C$ as the pair $(O',p')$ where $O'$ is the unique subnet of $O$ whose set of nodes is $\{x \mid x \notin C \cup \preset{C} \land \forall y \in C: \neg(x \# y)\}$ and $p'$ is the restriction of $p$ to the nodes of $O'$.
\end{definition}
\vspace{-0.5em}
Figure~\ref{fig:unfconf} highlights the configuration of $\{\acmakebook, \acsubmitpay, \acsubmitenroll\}$ is marked in green, with its corresponding cut being marked in blue. \ad{The notion $\uparrow\!\!C$ denotes the future part of $\beta$ that is possible after $C$, excluding anything that has already occurred or is in conflict with $C$.}

\vspace{-0.5em}
\begin{definition}[Extension]
Given a configuration $C$, its extension $C \oplus E$ is defined as $C \cup E$ while $C \cap E = \emptyset$.
\end{definition}
\vspace{-0.5em}

In Figure~\ref{fig:unfconf}, a possible extension of the marked configuration is the event representing the silent transition whose preset is the set of conditions marked in blue. The event marked in red cannot be added to this configuration, as it would violate conflict-freeness.

\vspace{-0.5em}
\begin{definition}[Configuration Isomorphism]
Let $C_1$ and $C_2$ be two configurations that lead to the same marking, i.e. $\mi{Mark}(C_1) = \mi{Mark}(C_2)$, then $I^2_1$ is the isomorphism of $\uparrow C_1$ and $\uparrow C_2$. This isomorphism induces a mapping from the finite extensions of $C_1$ onto the finite extensions of $C_2$: it maps $C_1 \oplus E$ onto $C_2 \oplus I^2_1(E)$.
\end{definition}
\vspace{-0.5em}

\vspace{-0.5em}
\begin{definition}[Local configuration]
Let $\beta= (B, E, G, l)$ be a branching process of the model $\mi{sp}$. The \term{local configuration} of an event $e \in E$, denoted by~$[e]$, is the set of events $e'$ such that $e' \leq e$. 
\end{definition}
\vspace{-1em}
\noindent \ad{For example, the local configuration $[\acsubmitpay] = \{\acmakebook, \acsubmitpay, t_1\}$ in Fig.~\ref{fig:unfconf}.}


\begin{definition}[Local configuration cost] Given a branching process $\beta = (B,E,F,p)$ and its associated net $N=(P,T,F,i,f)$, we define a cost function $Z([e],\zeta)$ for a local configuration $[e]$ as follows: \ch{$\sum_{e \in [e]} \zeta(p(e))$}, where $\zeta(t)$ assigns a numerical cost to each $t \in T$.
\end{definition}

\vspace{-1em}
\begin{definition}[Adequate order] A partial order $\prec$ on the finite configurations of the unfolding of a net $N$ is an \textit{adequate order} if:
\vspace{-0.5em}
\begin{itemize}
  \item $ \prec $ is well-founded, 
  \item $C_1 \subset C_2$ implies $C_1 \prec C_2$, and
  \item $ \prec $ is preserved by finite extensions; if $C_1 \prec C_2$ and $\mi{Mark}(C_1) = \mi{Mark}(C_2)$, then the isomorphism $I^2_1$ from above satisfies $C_1 \oplus E \prec C_2 \oplus I^2_1(E)$ for all finite extensions $C_1 \oplus E$ of $C_1$.
\end{itemize}
\end{definition}

\vspace{-1em}
\begin{definition}[Cut-off event] Let $\prec$ be an adequate order on the configurations of the unfolding of a net $N$. Let $\beta$ be a prefix of the unfolding containing an event $e$. The event $e$ is a \textit{cut-off event} of $\beta$ (with respect to $\prec$) if $\beta$ contains a local configuration $[e']$ such that
    (1) $\mi{Mark}([e]) = \mi{Mark}([e'])$, and
    (2) $[e'] \prec [e]$.
\end{definition}
\vspace{-0.5em}

As previously stated, \ch{the unfolding of a Petri net can be infinite due to loops or unbounded behavior. To address this, cut-off events are used to terminate parts of the unfolding when a previously encountered marking is reached again. This avoids redundant exploration of equivalent behavior. By applying this mechanism, one can construct a \emph{finite complete prefix}, which is a finite branching process that still captures all reachable markings of the original Petri net, if it is bounded.}


%

\section{Approach}\label{sec:approach}

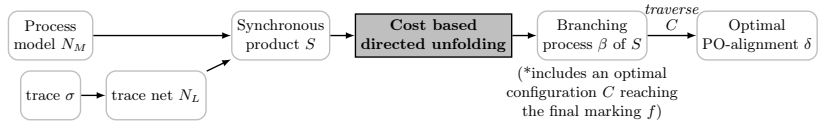
\begin{figure}[tb]
    \centering
    \resizebox{0.9\textwidth}{!}{
    \begin{tikzpicture}[
    transition/.style={rectangle, draw=lightgray, thick, minimum size=10mm, rounded corners=2mm},
    sync/.style={rectangle, draw=black, fill=lightgray, thick, minimum width=3mm, minimum height=8mm},
    node distance=0.2cm and 0.5cm, 
    >=latex 
]

\node (o1) [transition, align=center] {Process\\model $N_M$};
\node (o2) [transition, below=of o1] {trace $\sigma$};
\node (o3) [transition, right=of o2, align=center] {trace net $N_L$};
\node (o4) [transition, above right=of o3, align=center] {Synchronous\\product $S$};
\node (o5) [sync, right=of o4, align=center] {\textbf{Cost based}\\\textbf{directed unfolding}};
\node (o6) [transition, right=of o5, align=center] {Branching\\process $\beta$ of $S$};
\node[below=0.001cm of o6, align=center] {(*includes an optimal\\configuration $C$ reaching\\the final marking $f$)};
\node (o7) [transition, right=1cm of o6, align=center] {Optimal\\PO-alignment $\delta$};









\draw[->, thick] (o1) -- (o4);
\draw[->, thick] (o2) -- (o3);
\draw[->, thick] (o3) -- (o4);
\draw[->, thick] (o4) -- (o5);
\draw[->, thick] (o5) -- (o6);
\draw[->, thick] (o6) -- node[midway, above] {\shortstack{\emph{traverse} \\ $C$}} (o7);


\end{tikzpicture}
    }
    \caption{An overview of the unfolding alignment approach}
    \label{fig:enter-label}
\end{figure}

This section presents our approach for computing partial-order alignments using net unfolding. As no prior work has tackled alignments via Petri net unfoldings, we introduce an algorithm for this task and propose a few optimizations to enhance performance. Figure~\ref{fig:enter-label} shows an overview of the approach. 

\subsection{The Directed-Unfolding Algorithm}

In essence, the algorithm operates as follows. It begins by constructing the synchronous product $\mi{sp}= (P, T, F, \mi{im}, \mi{fm})$ of the model net $N_m$ and the event net $N_l$. It then unfolds $\mi{sp}$ in a directed, cost-based manner to create a branching process $\beta = (B,E,G,p)$, terminating once the final marking $\mi{fm}$ is reached in some configuration, i.e., $\exists C \in \beta$ such that $\mathit{Mark}(C) = \mi{fm}$. To simplify this check, we added a dummy end event to the synchronous product, allowing us to use the local configuration of this event instead. Thus, the stopping condition becomes: $\exists x \in \beta$ such that $\mathit{Mark}([x]) = \mi{fm}$. The corresponding configuration $[x]$ defines the optimal partial-order alignment $\delta$. With this idea, Algorithm~\ref{alg:basic} presents the pseudocode for computing an optimal partial-order alignment via directed unfolding, built on the idea of cut-off events and an adequate order~\cite{esparza1996improvement}.

\begin{center}
\scalebox{0.9}{  
\begin{algorithm}[H]
\caption{A pseudocode algorithm for computing partial-order alignments through directed unfolding}
\label{alg:basic}
\KwData{A net $N_m$ with a reachabile final marking, an event net $N_l$}
\KwResult{An optimal partial-order alignment $\delta$}
$cost \gets
      \zeta(t) =
        \begin{cases}
          0      & \text{if $t$ is synchronous}\\
          1      & \text{if $t$ is model or log}\\
          0.0001 & \text{if $t$ is silent}
        \end{cases}       
$\;    
$sp = (P, T, F, \mi{im}, \mi{fm}) \gets N_l \times N_m$ {\color{gray}\Comment*[r]{Create a synchronous product $sp$, with its initial marking $\mi{im}$ and final marking $\mi{fm}$}}
$\beta \gets \mathit{IntializeBranchingProcess}(\mi{sp}, \mi{im})$\;
$pe \gets \mathit{PossibleExtensions}(\beta)$\;
$\mathit{cutoff} \gets \emptyset$\;
$\delta \gets \emptyset$\;
\While{$\mathit{pe} \neq \emptyset$}{
choose an event $e = (t, X)$ in $pe$ such that $Z([e], \zeta)$ is minimal\;
  \If{$[e] \cap \mi{cutoff} = \emptyset$}{
    $\beta \gets \beta \cup e$\;
    \ch{$\beta \gets \beta \cup \{ (s', e)  \mid s \in \postset{p(e)}\}, p(s') \gets s$}{\color{gray}\Comment*[r]{for each place $s$ in the postset of the transition corresponding to $e$, create a condition $s'$ and add it to the branching process}} 
    \If{$\exists x \in \beta; \mathit{Mark}([x]) = \mi{fm}$ {\color{gray}\Comment*[r]{Reached the final marking}}}{
        $\delta \gets [x]$ {\color{gray}\Comment*[r]{Update optimal alignment}}
        \ch{\textbf{BREAK} the while loop.}
    }
    $pe \gets \mathit{PossibleExtensions}(\beta)$\;
    \If{$e$ is a cut-off event of $\beta$}{
        $\mathit{cutoff} \gets \mathit{cutoff} \cup \{e\}$\;
    }
  }{
    $pe \gets pe \setminus e$\;
  }
}
\ch{\Return $\delta$ \Comment*[r]{Return this optimal alignment}}
\end{algorithm}
    }
\end{center}
We begin by computing $\mi{sp}$ and initializing $\beta$ using the initial marking $\mi{im}$ of $\mi{sp}$, adding a \emph{condition} to $\beta$ for each place in $\mi{im}$ (see Lines~1-3). 
A \textit{condition} $(s, e)$ consists of a place in $\mi{sp}$ and a pointer to its input event, while an \textit{event} $(t, X)$ consists of a transition $t$ in $\mi{sp}$ and a set $X$ of input conditions. In pseudocode, conditions are represented as $(s, e)$ or $(s, \emptyset)$, and events as $(t, X)$.






After initializing the branching process $\beta$, we begin unfolding $\mi{sp}$ by firing enabled transitions (as possible extensions, see Line~4), adding each as a new event in $\beta$, and appending its post-set as new conditions. The transition selection must follow the adequate order to ensure correctness and optimality during directed unfolding.

\begin{figure}[tp]
\includegraphics[width=\textwidth]{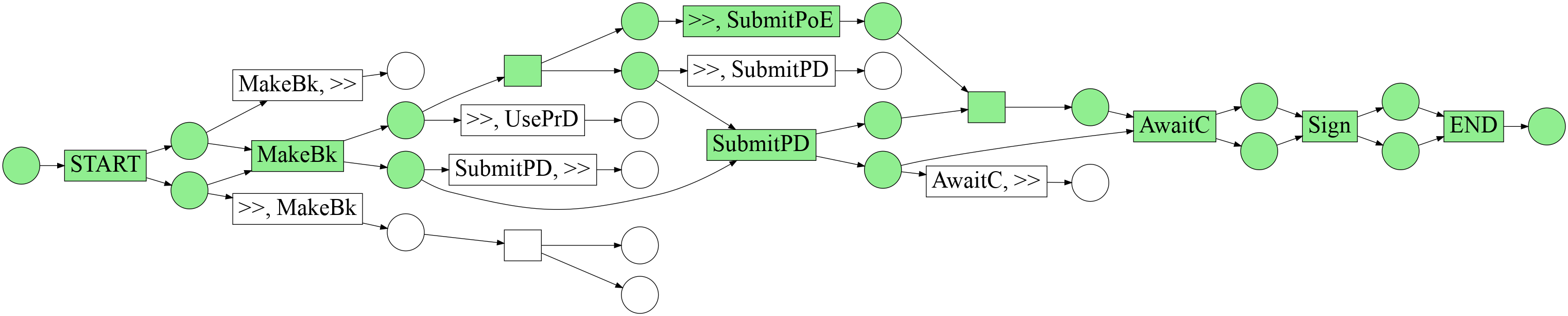}
\caption{\ch{The branching process constructed by Algorithm \ref{alg:basic} on the synchronous product in \ref{fig:sp}. The optimal partial-order alignment is highlighted in green.}}
\label{fig:bplabeled}
\end{figure}

\begin{figure}[tp]
\includegraphics[width=\textwidth]{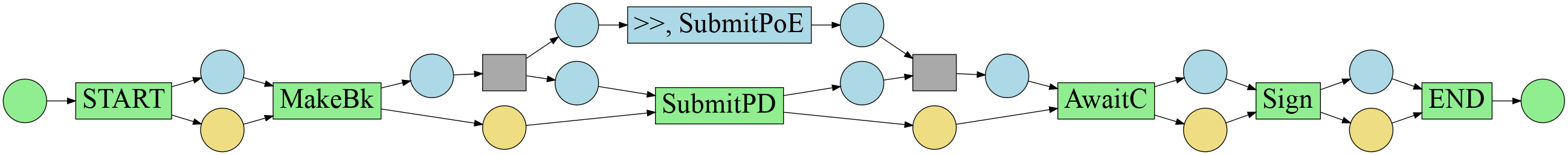}
\caption{\ad{The optimal partial-order alignment as an occurrence net.}}
\label{fig:optimalpoalignment}
\end{figure}

\ch{We define the adequate order $\prec$ using the cost function $Z([e], \zeta)$ of a local configuration $[e]$, based on a predefined cost function $\zeta$. \ch{We use the standard cost function for $\zeta$, as listed in Algorithm~\ref{alg:basic}.} \ad{Thus, $e \prec e'$ if and only if $Z([e], \zeta) \leq Z([e'], \zeta)$}.
}
To break any ties and maintain the adequate order $\prec$, each event
$e$ is assigned a unique numerical identifier using Python’s built-in $\mi{id}()$ function\footnote{\scriptsize\url{https://docs.python.org/3/library/functions.html\#id}}, i.e., \ad{if $Z([e], \zeta) = Z([e'], \zeta)$, we order the events by $\mi{id}(e) < \mi{id}(e')$.}

We iteratively append the \emph{best} possible extension $e$ to $\beta$ (see Lines~7-11), check whether the final marking is represented (see Lines~12-15), and, if not, continue with computing new possible extensions and the next \emph{best} extension (see Lines~16-20). Once the final marking is reached, we return the corresponding configuration, which defines an optimal alignment guaranteed by $\prec$. Figure~\ref{fig:bplabeled} illustrates a directed unfolding of $\mi{sp}$ from Fig.~\ref{fig:sp}, and Figure~\ref{fig:optimalpoalignment} shows the resulting partial-order alignment.

    

\ad{
\vspace{-0.5em}
\begin{theorem}[Correctness of Unfolding Alignment]
Given a Petri net $N_m$ and an event net $N_l$, let $\delta$ be the alignment returned by Algorithm~\ref{alg:basic}, which is defined by an optimal configuration $[e]$ in the branching process $\beta$ of the synchronous product $sp = N_m \times N_l$. Then, $\delta$ is a valid, optimal partial-order alignment. Specifically:
    (1) $\delta \subseteq \beta$ corresponds to a valid firing sequence in $sp$ from its initial marking $\mi{im}$ to final marking $\mi{fm}$.
    (2) If a valid alignment exists, the unfolding contains it, and the algorithm will find it.
    (3) The alignment $\delta$ minimizes the cost $Z([e], \zeta)$ among all valid alignments.
    (4) The algorithm terminates.
\end{theorem}
\vspace{-1em}
\begin{proof}
For (1), all added events correspond to enabled transitions, preserving soundness (guaranteed by the definition of a branching process). For (2) and (4), by Theorem 4.3.4 and Lemma 4.4.1 in~\cite{adriansyah2014aligning}, the final marking $\mi{fm}$ is reachable in $sp$, thus, also reachable in $\beta$. Once it is reached, our algorithm terminates. For (3), the ERV unfolding algorithm guarantees that $\beta$ contains at least one path to $\mi{fm}$. Our algorithm explores configurations in order of increasing cost guaranteed by $\prec$ and selects the first reaching $\mi{fm}$, ensuring optimality. 
\end{proof}
\vspace{-0.5em}
}

To enhance performance, we implement a few optimizations. The set of possible extensions ($pe$ in Algorithm \ref{alg:basic}) is implemented as a priority queue, ensuring efficient selection of the lowest-cost extensions. Moreover, we store events indexed by their marking, which simplifies cut-off event detection. 
Additionally, each event retains its local configuration and cost, enabling faster cost computations for new extensions and quick identification of cut-off events. We refer to this algorithm as \foldan. 

\subsection{Using Heuristics}
\ad{In the previous section, we defined the adequate order $\prec$ using the cost function $Z([e], \zeta)$, following a Dijkstra approach. To reduce backtracking, we present a heuristic function $h_{\mi{sp}}(\mi{Mark}([e]))$, which estimates a lower bound on the remaining cost for an extension $e$ to reach the final marking $\mi{fm}$, adopting an A* approach. This heuristic is computed by solving the marking equation of the synchronous product for minimum cost, as proposed in~\cite{adriansyah2014aligning}. The new adequate order $\prec$ is then defined using $Z([e], \zeta) + h_{\mi{sp}}(\mi{Mark}([e]))$. We omit the subscript $sp$ when the context is clear. 
While Theorems 4.4.8 and 4.4.9 in~\cite{adriansyah2014aligning} prove that this heuristic yields a valid lower bound for traditional alignment computation, its permissibility in directed unfolding has not been established~\cite{bonet2008directed}. We now prove that the heuristic remains valid in our unfolding-based setting.}
%
We first show that the computed heuristic value there corresponds with a certain event $e$ in $\beta$.

\vspace{-0.5em}
\begin{theorem}
Given an easy sound Petri net $N$ and \ch{its reachability graph $\mi{RG}(N)$}, a complete, finite prefix $\beta = (B, E, G, p)$ of $N$, and the marking-equation based heuristic function $h$, then $\forall e \in E, \exists s \in \mi{RG}(N): h(\mathit{Mark}([e])) = h(s)$.
\end{theorem}
\vspace{-0.5em}
\noindent \ch{This follows trivially from the definition of a branching process. Since each $[e]$ represents a valid firing sequence leading to a reachable marking $s$, and $h$ depends solely on the marking, the heuristic values must match.}




Since the heuristic $h(m)$ estimates the minimal cost from a marking $m$ to $\mi{fm}$, it preserves alignment optimality. \ch{We update Line~8 of Algorithm~\ref{alg:basic} to select the next event $e$ that minimize $\min_{e\in\mi{pe}}Z([e], \zeta) + h(\mi{Mark}([e])$.} Now, we show that the adequate order $\prec$ is preserved~\cite{bonet2008directed}.
The first condition ($\prec$ is well-founded) holds trivially, since cost and heuristic values are non-negative, ensuring a minimal cost always exists. The third condition ($\prec$ is preserved by finite extensions) is also straightforward. Since the heuristic value for any two extensions with the same marking is the same, the adequate order remains unchanged.
%
%
We now prove the second condition, namely that $C_1 \subset C_2$ implies $C_1 \prec C_2$.

\ch{
\vspace{-0.5em}
\begin{theorem}
    Let $e_1$ and $e_2$ be two events such that $[e_1] \subset [e_2]$, with a cost function $Z([e],\zeta)$, and the heuristic function $h$. Then, $[e_1] \prec [e_2]$, i.e., $Z([e_1],\zeta) + h(\mathit{Mark}([e_1])) \leq Z([e_2], \zeta) + h(\mathit{Mark}([e_2]))$ holds.
\end{theorem}
\vspace{-0.8em}
\begin{proof}
    Assume, for contradiction, that $Z([e_1],\zeta) + h(\mathit{Mark}([e_1])) > Z([e_2], \zeta) + h(\mathit{Mark}([e_2]))$. Let $E_d = [e_2]\setminus [e_1]$, i.e., $[e_1] \cup E_d = [e_2]$. This implies that 
    $h(\mathit{Mark}([e_1])) > Z(E_d, \zeta) + h(\mi{Mark}([e_2])$. 
    This implies that there exists a path from $\mathit{Mark}([e_1])$ to the final marking $\mathit{fm}$ via $E_d$ and $\mathit{Mark}([e_2])$ that is cheaper than the heuristic from $\mathit{Mark}([e_1])$ alone. This contradicts the assumption that $h$ is a lower bound. Therefore, the inequality must hold, and thus $[e_1] \prec [e_2]$.
\end{proof}
}
\vspace{-0.5em}

\ch{Thus, we have shown that the heuristic function is valid for use in directed unfoldings and that $\prec$ holds. We refer to this heuristic-enhanced version as \foldah. We compare \foldan and \foldah in our evaluation.}

\FloatBarrier



\section{Evaluation}\label{sec:evaluation}
This section outlines the evaluation setup for the proposed approach. 
%
\ad{We first examine the feasibility of our approach.} Then,
Experiment~1 examines the impact of \emph{model structure} and the \emph{placement of deviations} on unfolding alignment performance. We investigate three model constructs, namely concurrency, choice, and loops. 
Experiment~2 evaluates unfolding alignment performance against \dijk and \astar based alignments using 7 real-world and 6 benchmark logs. 

To assess performance, we measure the following metrics for each alignment: 
    elapsed time (ET),
    number of queued states ($\#$QS), and
    number of visited states ($\#$VS).
To determine how certain attributes of the input data affect performance, we also measure 
    cost,
    trace length, and
    number of transitions in the synchronous product ($\#$SPT).

All experiments were performed on a Lenovo IdeaPad 5 equipped with an AMD Ryzen 7 5700U CPU with a clock speed of 1.8 GHz (up to 4.3 GHz turbo) and 16GiB of memory. The code was executed using Python 3.12.0, on Ubuntu 22.04.4 LTS 64-bit. 
For the baseline techniques, we use PM4PY's implementations of Dijkstra (no heuristic) and \astar (with heuristic) alignments. Since our implementation is also built upon PM4Py, this allows us to control for programming language, preprocessing, and data structure differences. 
The source code is publicly available on GitHub\footnote{\url{https://github.com/DouweGeurtjens/unfolding-alignments}}~\cite{douwegeurtjens_2025_15552463}.


\mysubsubsection{Feasibility.} We began by testing the feasibility of our approach using the real-life healthcare dataset Sepsis (SP) \cite{sepsis} and a \emph{perfectly fitting} model discovered using the inductive miner in PM4PY. The log contains 1050 traces, with lengths ranging from 3 to 185 (avg. 14.49). 
We evaluated the \ch{\emph{naive} unfolding alignments (\foldan)}, i.e., without heuristics, against Dijkstra and A*.
All three techniques produced optimal alignments (identical costs). Interestingly, \foldan achieved a lower median elapsed time (0.023s) compared to Dijkstra (7.425s) and \astar (0.614s), along with fewer queued and visited states. However, we observe that \foldan has extreme outliers, with a maximum elapsed time of 2702 seconds, far exceeding \astar (25s) and Dijkstra (87s). This is also reflected in the mean of elapsed time, where \foldan is much higher than \astar (5.43s vs 0.69s).

\begin{table}[t]
\centering
\caption{A taxonomy of the artificial datasets used in Experiment 1.}
\label{tab:taxonomy}
\resizebox{\textwidth}{!}{%
\begin{tabular}{@{}lllllll@{}}
\toprule
Model type & Construct   & Breadth (Min-Max) & Depth (Min-Max) & Nesting Factor (Min-Max) & Nesting Breadth (Min-Max) & Nesting Depth (Min-Max) \\ \midrule
C            & Concurrency & 2-\toch{12}              & 1-15            & NA                       & NA                        & NA                      \\
E            & Choice      & 2-15              & 1-15            & NA                       & NA                        & NA                      \\
CN           & Concurrency & 2-2               & 1-5             & 1-5                      & 2-2                       & 1-5                     \\
EN           & Choice      & 2-2               & 1-5             & 1-5                      & 2-2                       & 1-5                     \\
L            & Loop        & 1-1               & 1-5             & NA                       & NA                        & NA                      \\ \bottomrule
\end{tabular}
}
\end{table}

\begin{figure}[t]
    \centering
    \begin{minipage}{0.48\textwidth} 
            \subfloat[A C-type model with a breadth of 3 and depth of 5]{\centering
            \includegraphics[width=\textwidth]{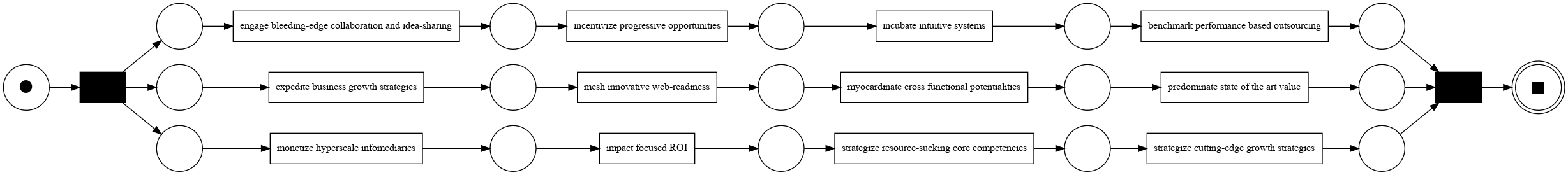}}
        \vspace{0.1em} 

            \subfloat[An L-type model with a breadth of 1 and depth of 5]{\centering
            \includegraphics[width=\textwidth]{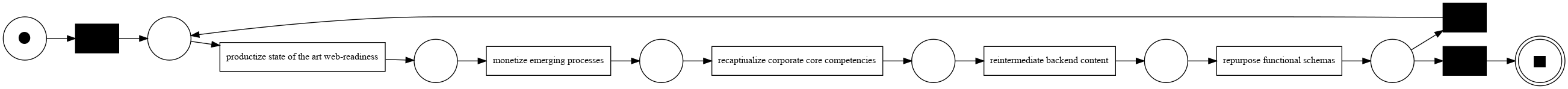}}

    \end{minipage}
    \hfill
    \begin{minipage}{0.48\textwidth} 

            \subfloat[A CN-type model with a breadth of 2, depth of 2, nesting factor of 3, nesting breadth of 2, and nesting depth of 2]{\centering
            \includegraphics[width=\textwidth]{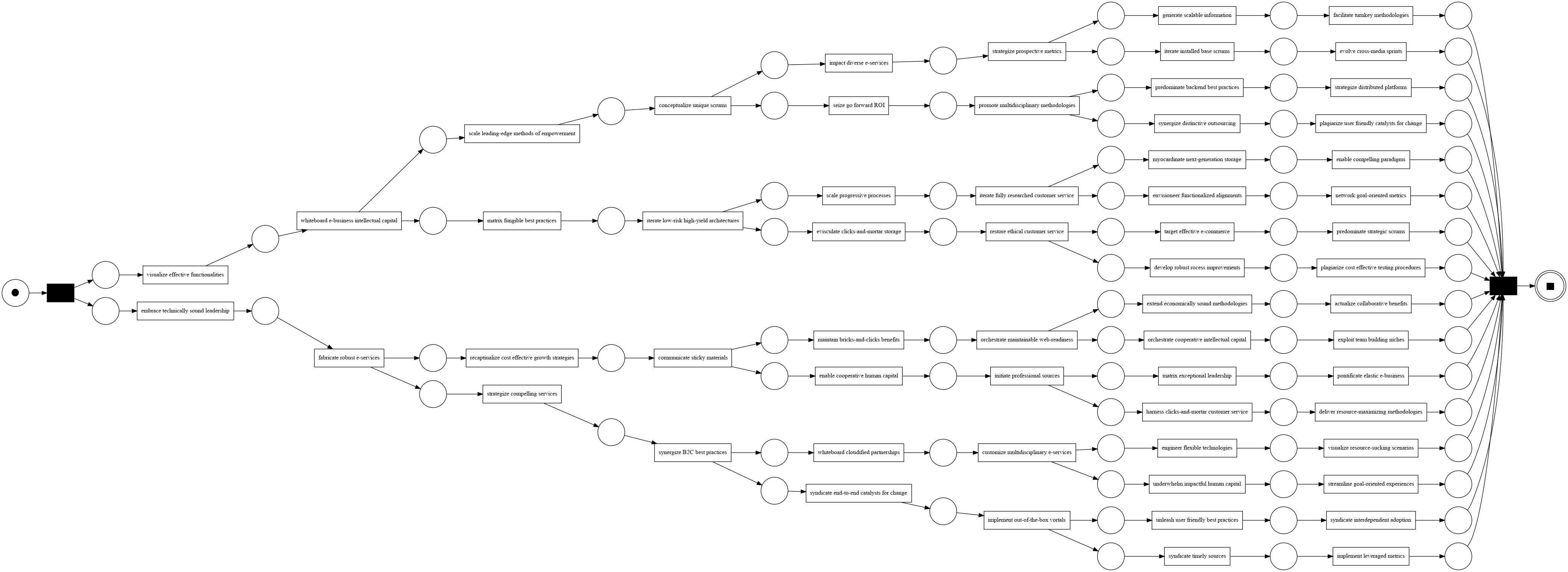}}

    \end{minipage}
    \caption{Examples of the artificially generated models used in experiment 1.}
    \label{fig:artimod}
\end{figure}

\subsection{Experiment 1 - Impact of Model Structure and Deviations}

\subsubsection{Setup} 
\ch{Here, we compare \foldan and \foldah.} To assess the impact of concurrency, choice, and loops, we created artificial process models that isolate each construct. Each model type is defined by breadth (number of branches), depth (branch length), and for concurrency/choice models, nestedness (how often a construct is repeated, which may exponentially increase model size). Fig.~\ref{fig:artimod} illustrates representative examples, and Table~\ref{tab:taxonomy} outlines parameter ranges. Each model was assigned unique transition labels, and 50 random traces were generated per model, resulting in 24,250 traces across 485 models.

To simulate deviations, we systematically removed one event per trace at the start, middle, or end. We decided not to insert or replace the events due to the combinatorial explosion of options. Each alignment computation was limited to 100 seconds per trace.

\begin{figure}[tb]
\centering
\includegraphics[width=0.24\textwidth]{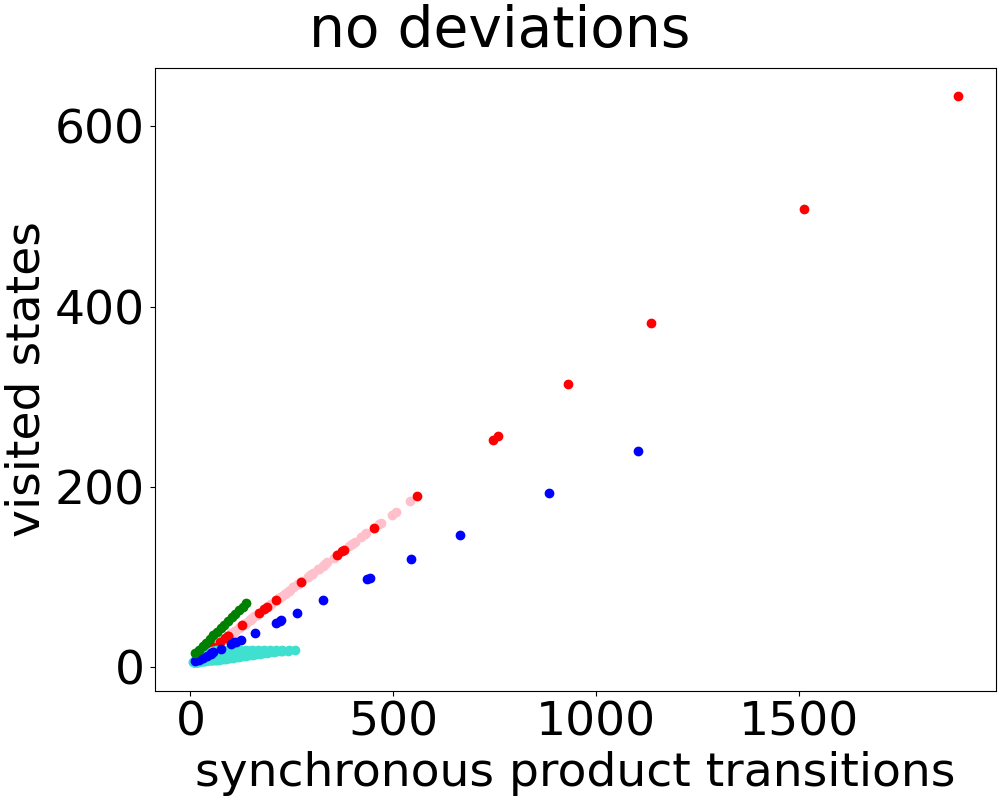}
\includegraphics[width=0.24\textwidth]{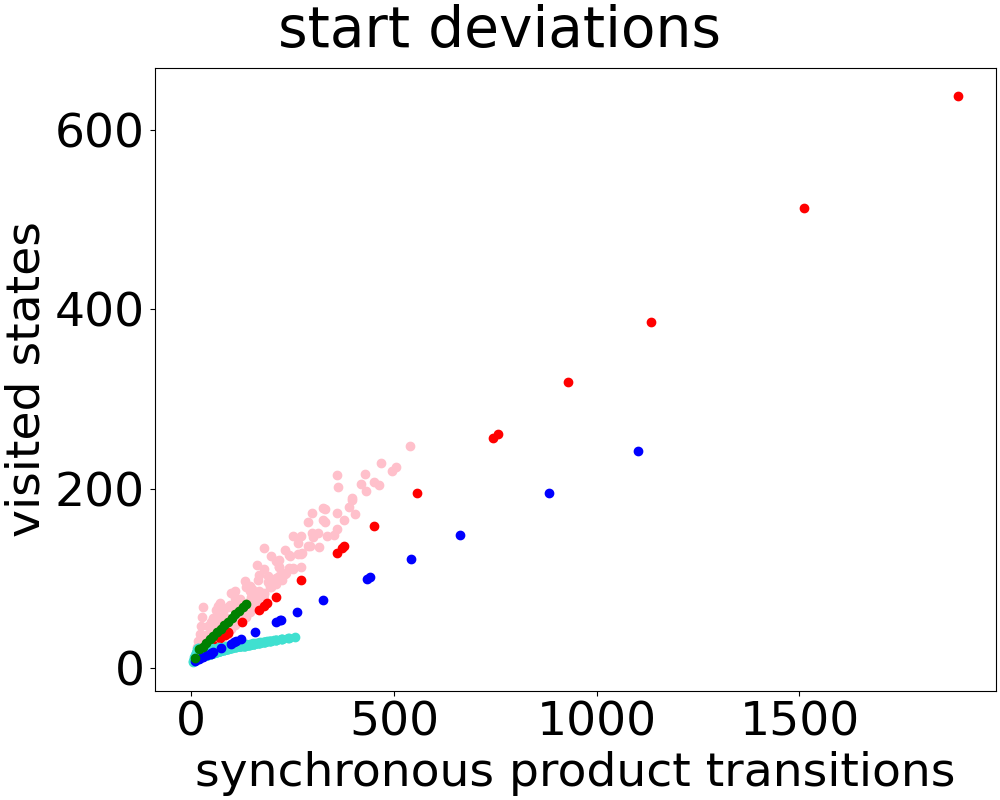}
\includegraphics[width=0.24\textwidth]{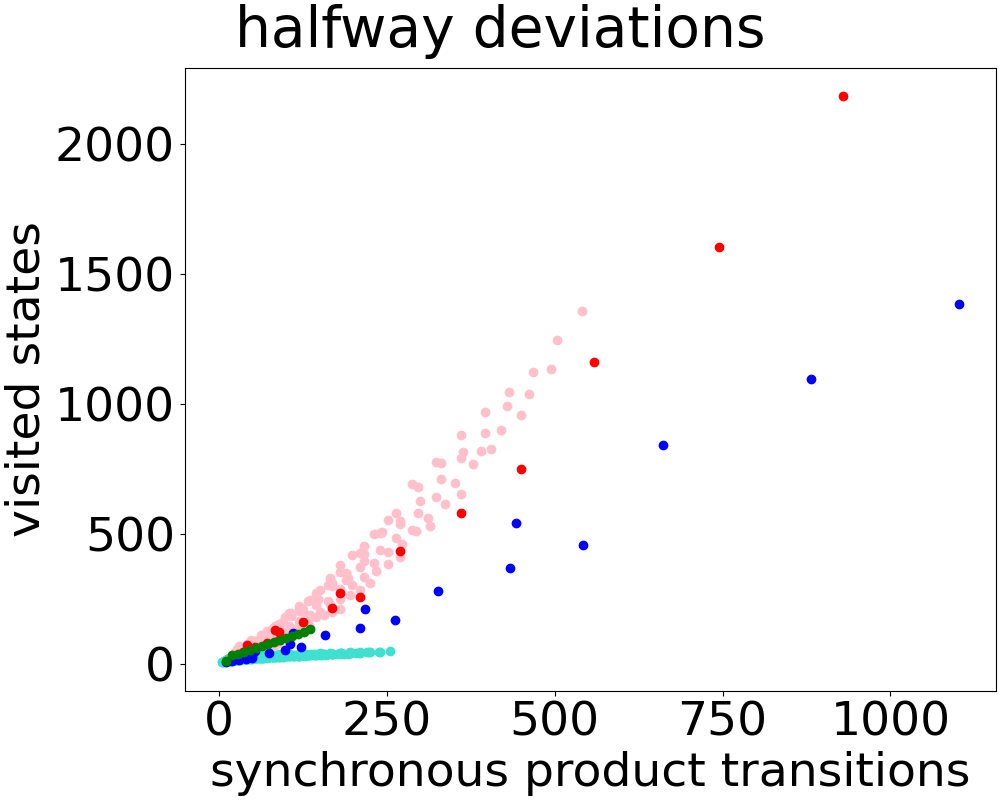}
\includegraphics[width=0.24\textwidth]{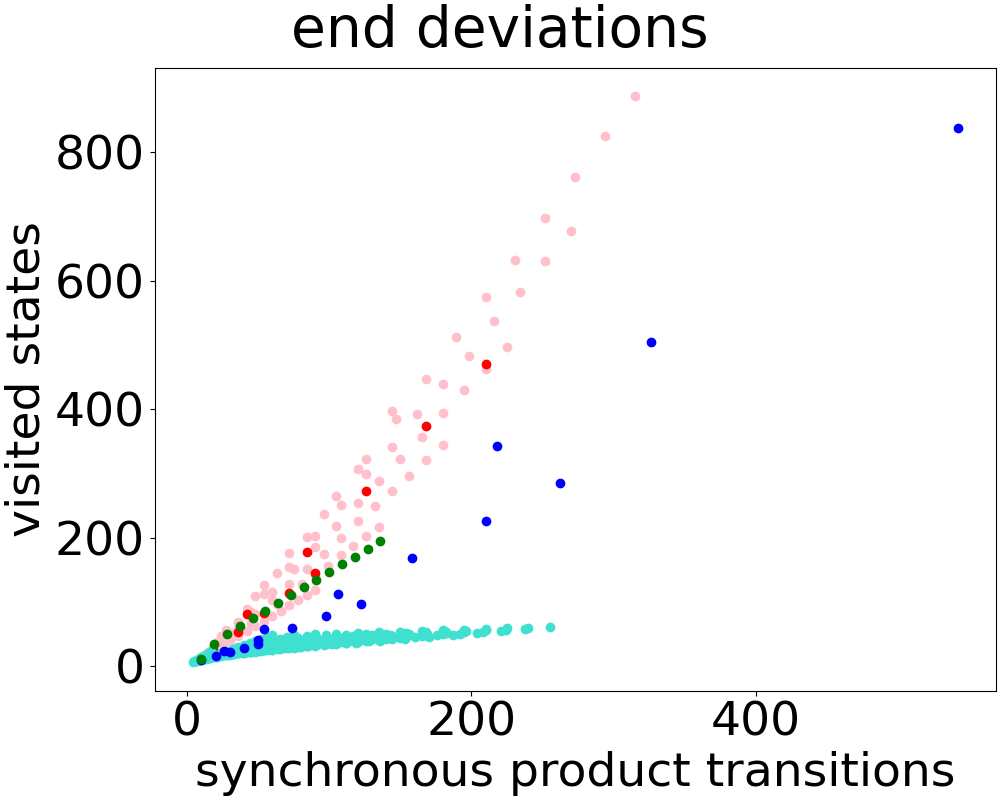}
\centering
\includegraphics[width=0.24\textwidth]{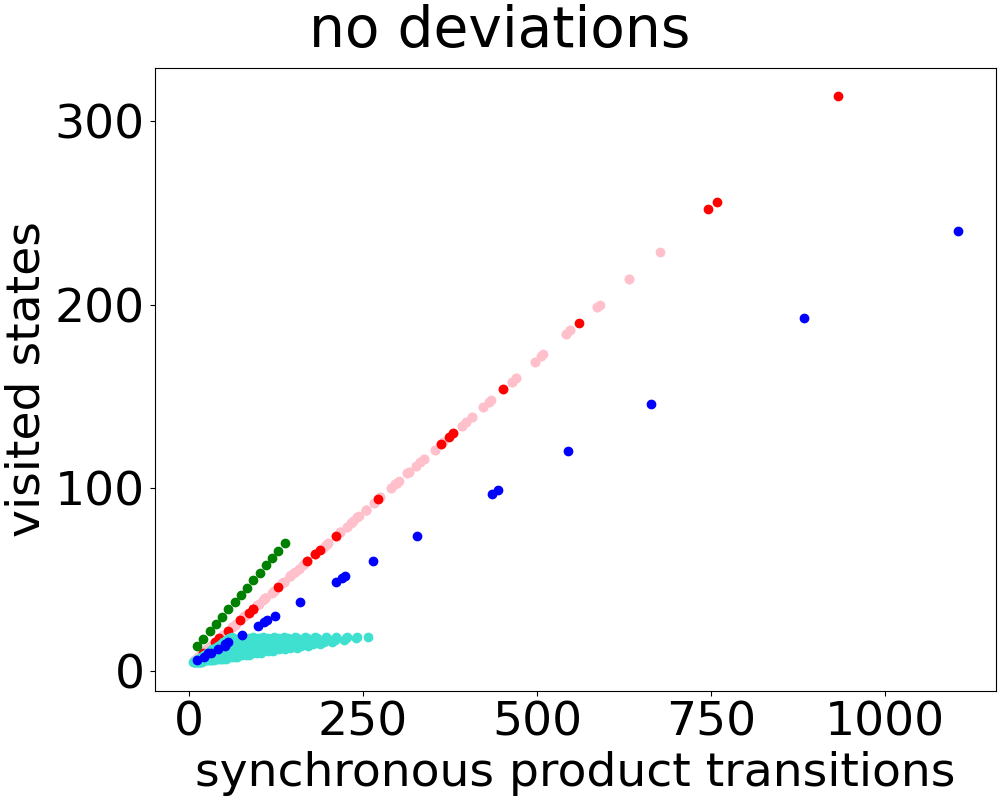}
\includegraphics[width=0.24\textwidth]{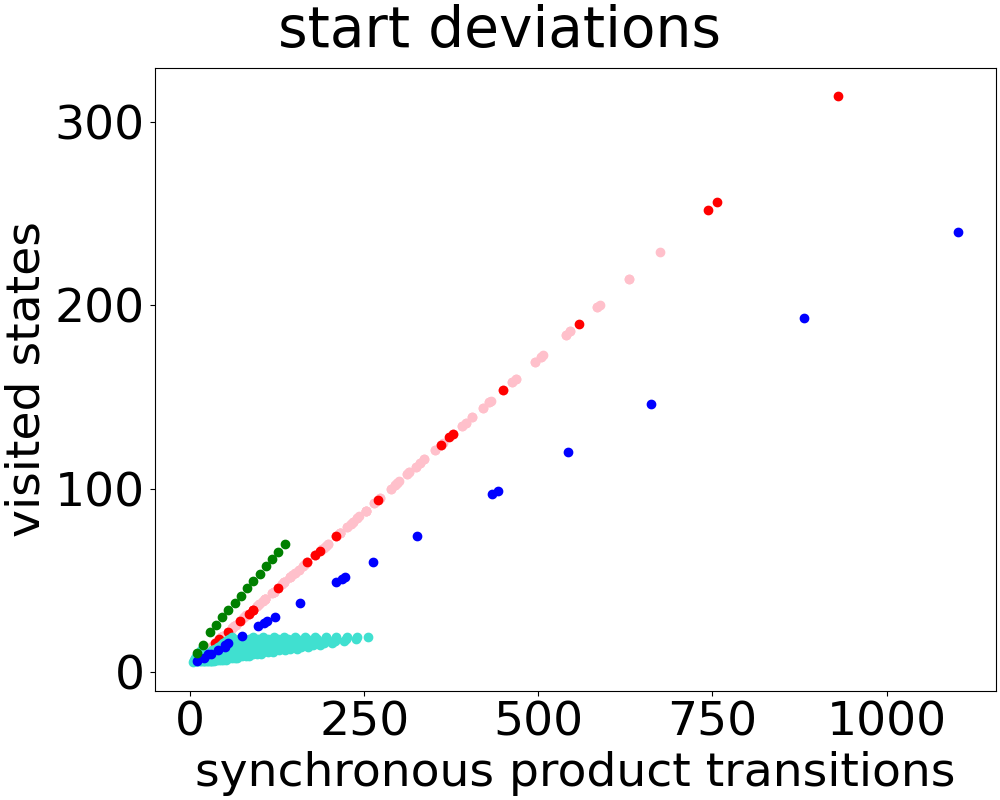}
\includegraphics[width=0.24\textwidth]{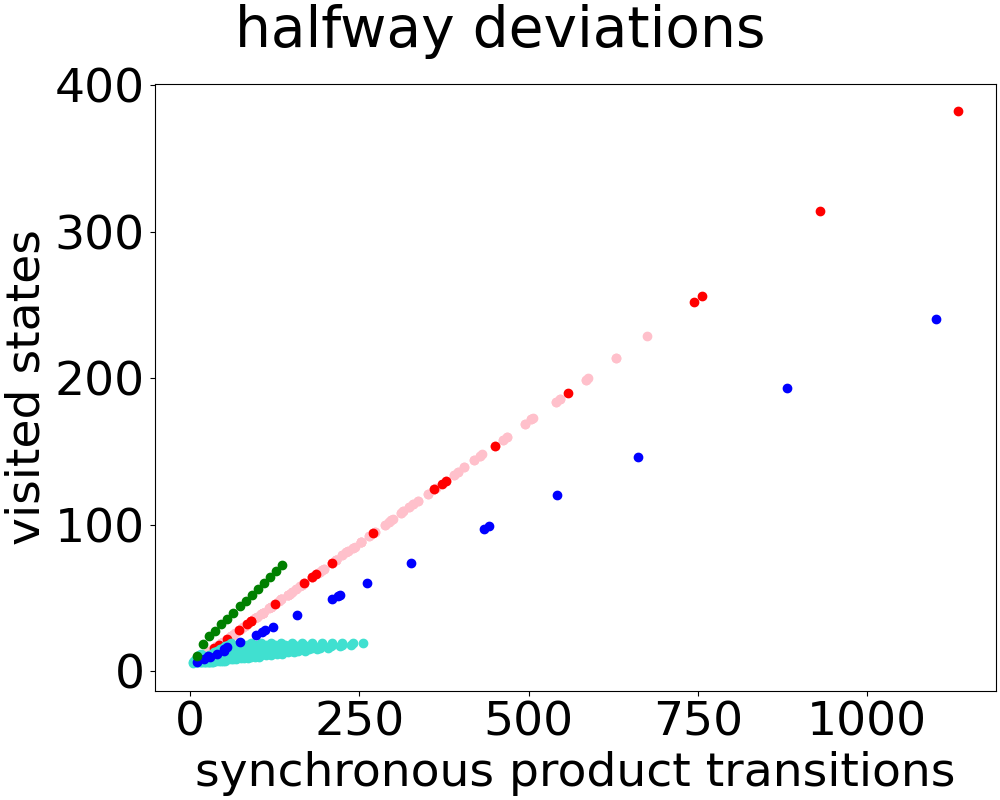}
\includegraphics[width=0.24\textwidth]{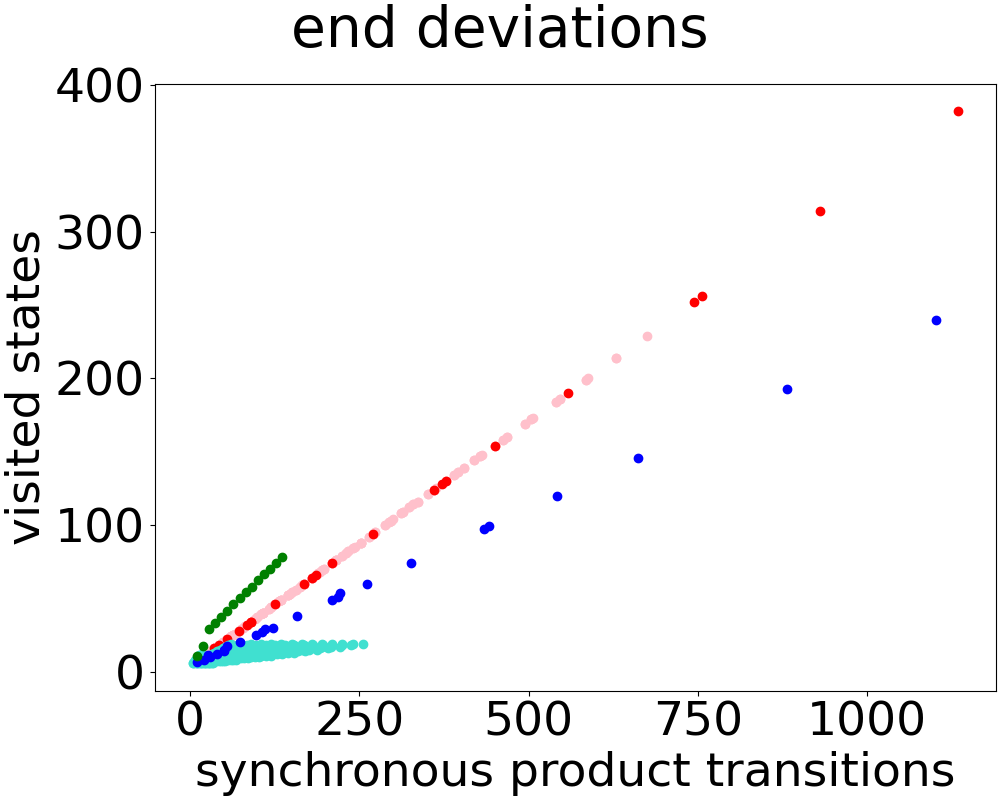}
\includegraphics[width=0.3\textwidth]{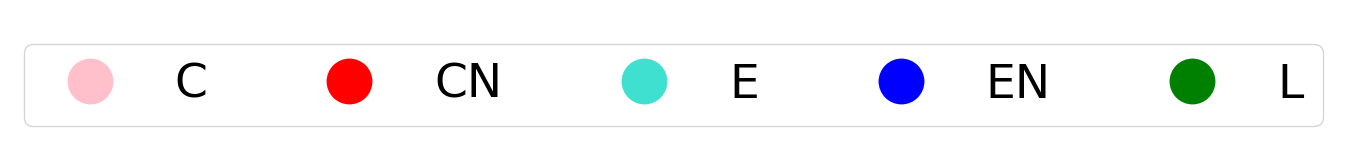}
\caption{$\#$VS vs $\#$SPT, across model types and deviation placements, comparing \foldan (top four) with \foldah (bottom four).}
\label{fig:sp_v_heuristic}
\end{figure}

\mysubsubsection{Results} 
Fig.~\ref{fig:sp_v_heuristic} displays scatterplots showing the relationship between synchronous product transitions and visited states. The relationship is generally linear across model types and deviation placements. For \foldan, coefficients vary with deviation placement, whereas for \foldah, coefficients remain stable.

\begin{figure}[tb]
\centering
\includegraphics[width=0.24\textwidth]{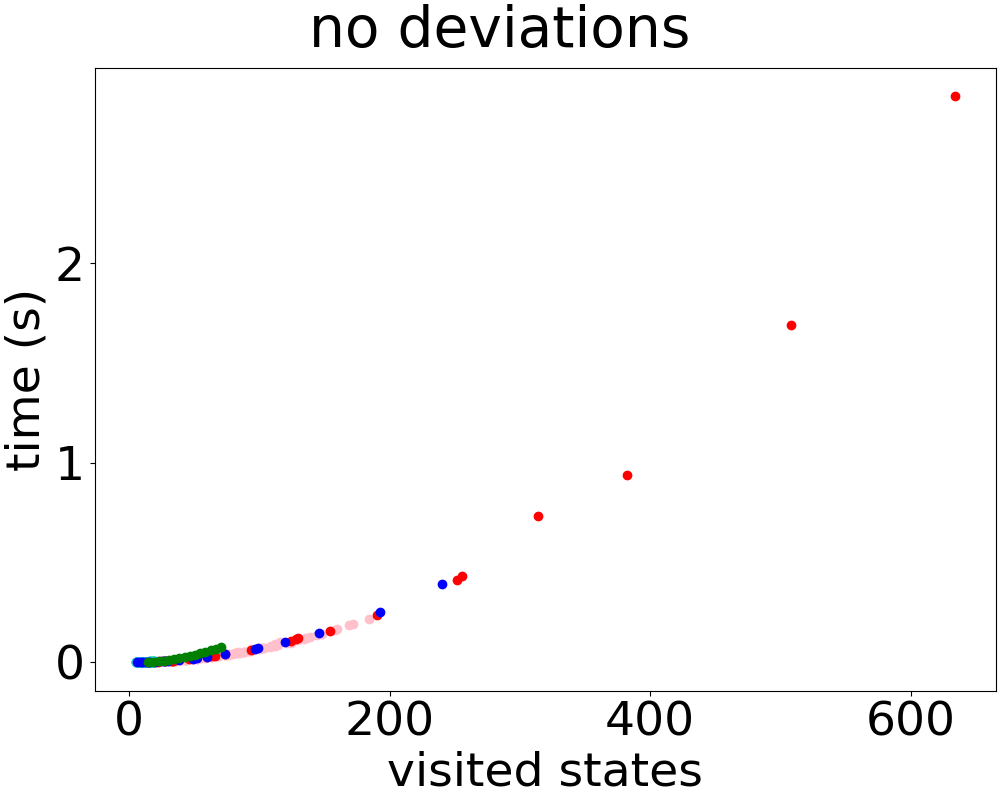}
\includegraphics[width=0.24\textwidth]{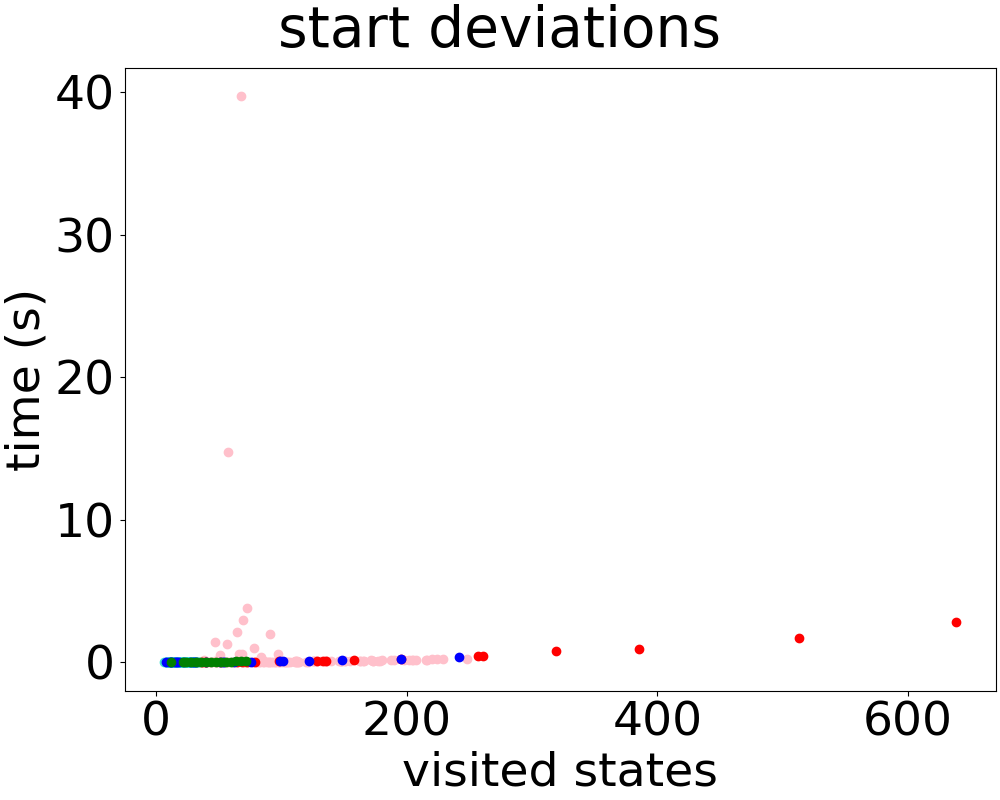}
\includegraphics[width=0.24\textwidth]{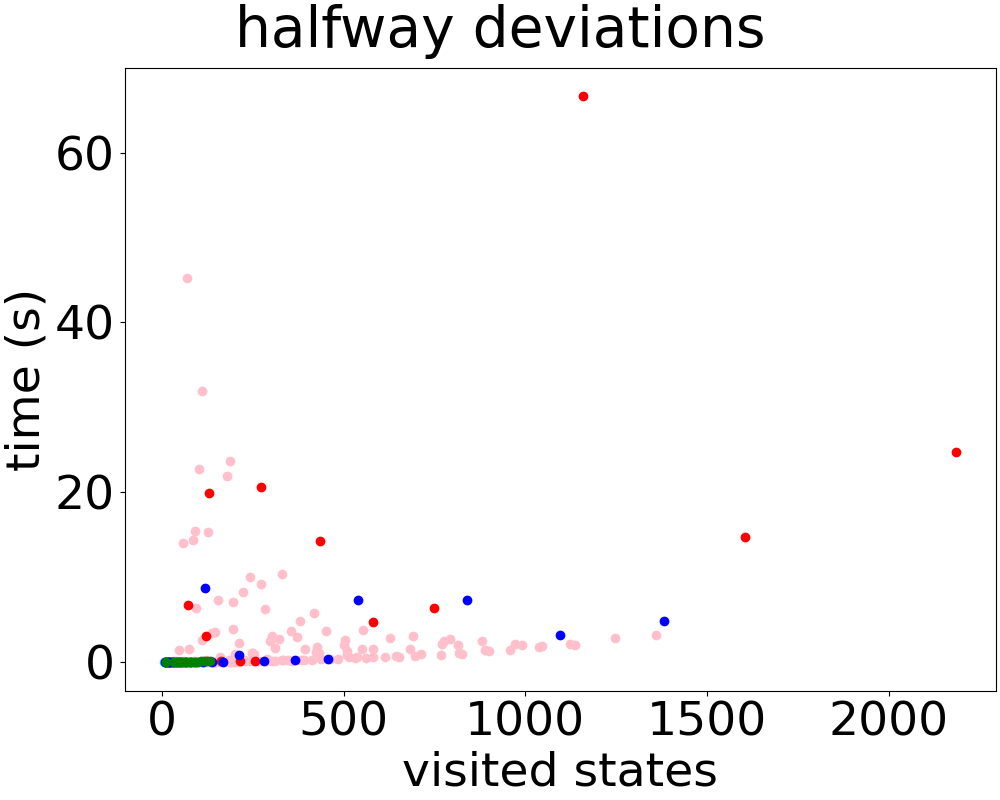}
\includegraphics[width=0.24\textwidth]{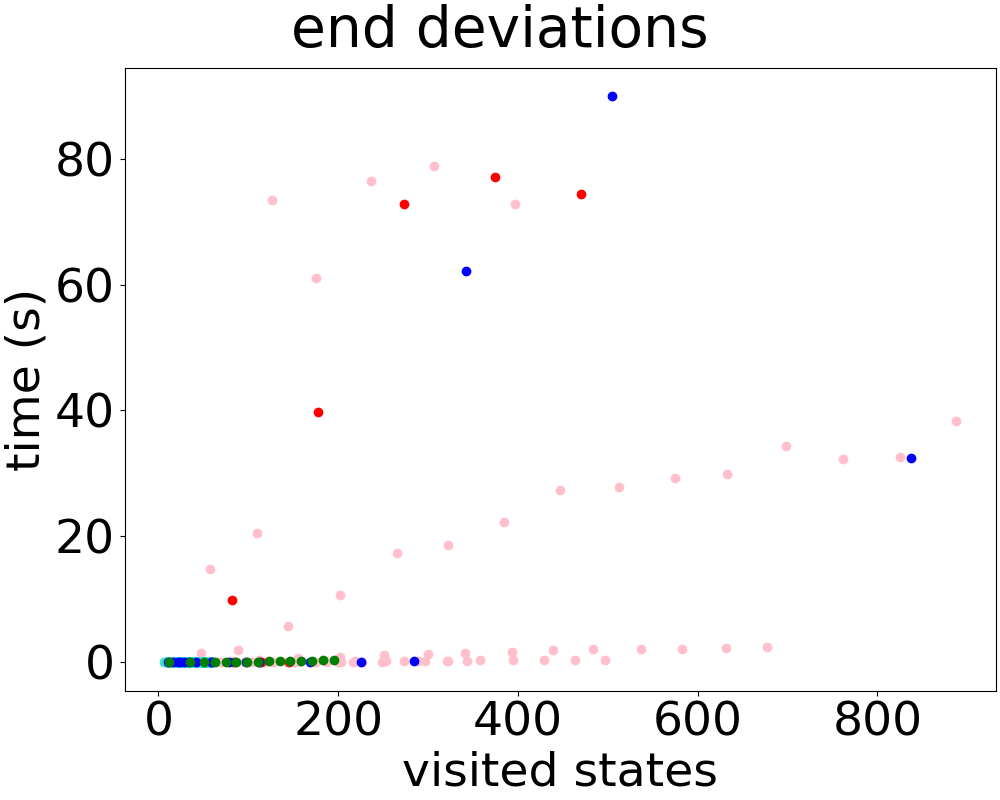}
\centering
\includegraphics[width=0.24\textwidth]{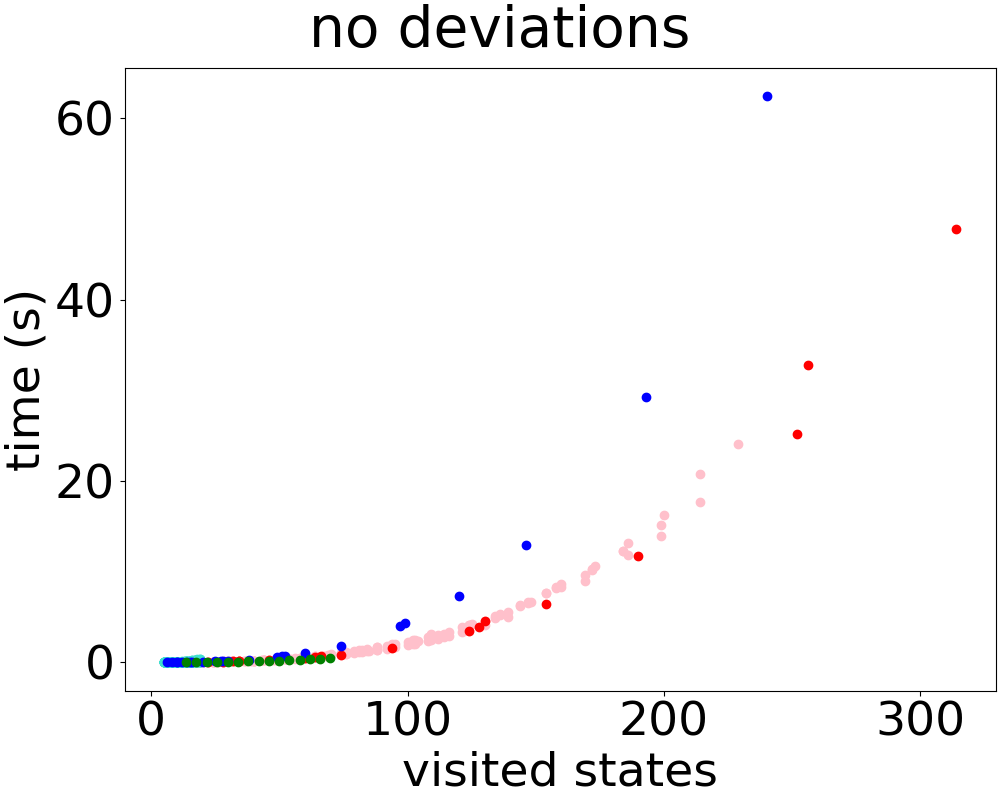}
\includegraphics[width=0.24\textwidth]{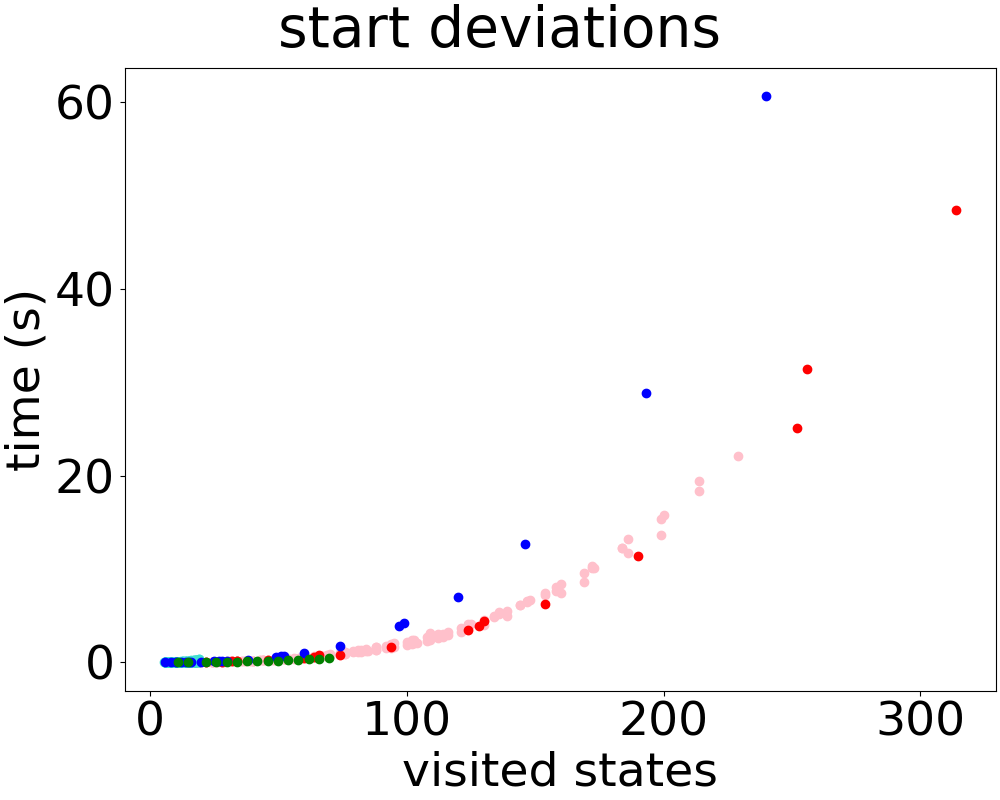}
\includegraphics[width=0.24\textwidth]{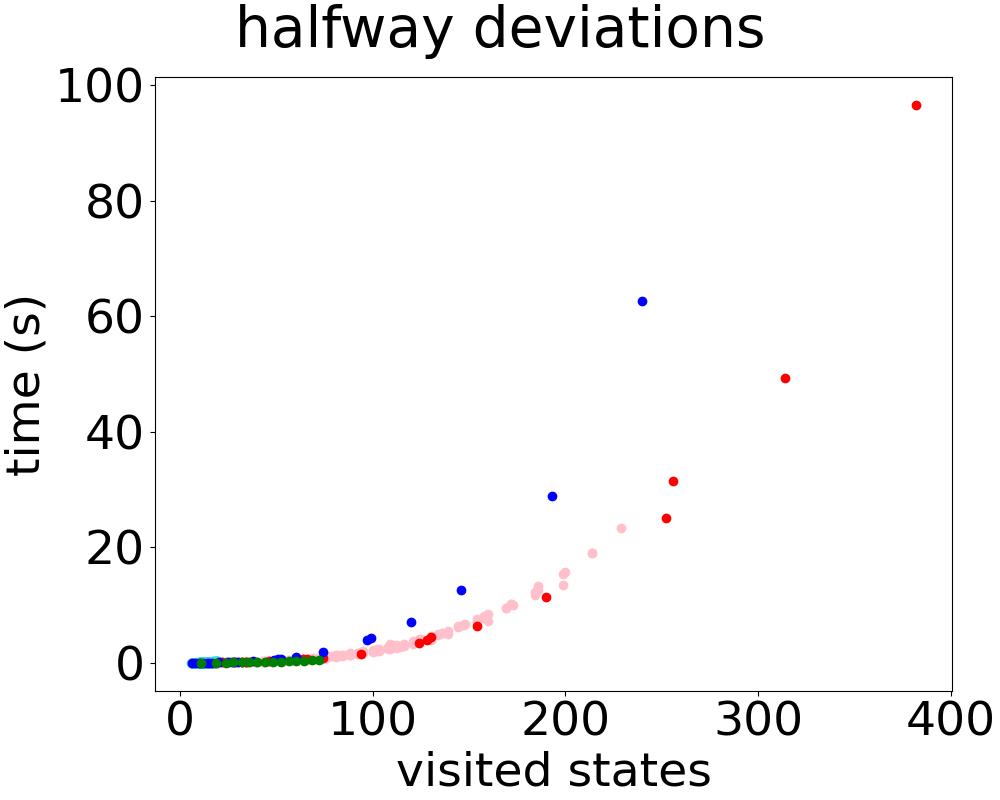}
\includegraphics[width=0.24\textwidth]{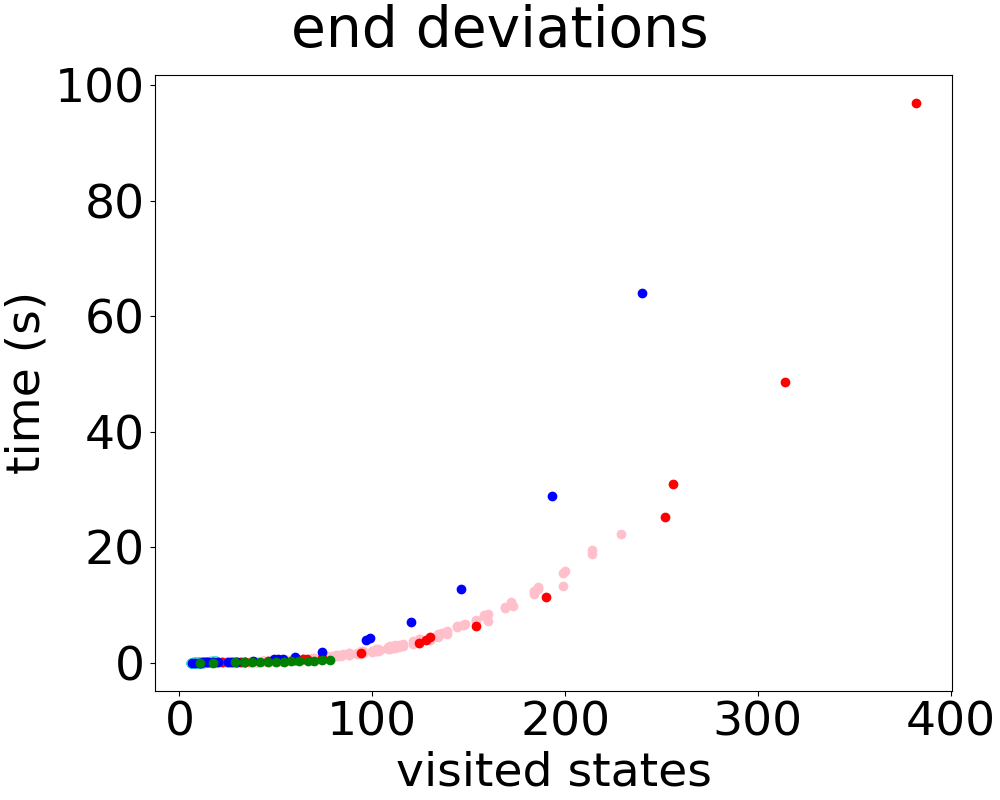}
\includegraphics[width=0.3\textwidth]{figures_34/stage_one/legend.png}
\caption{Elapsed time per $\#$VS across model types and deviation placements, comparing \foldan (top four) versus \foldah (bottom four).}
\label{fig:v_t_heuristic}
\end{figure}

Fig.~\ref{fig:v_t_heuristic} depicts visited states versus elapsed time. Without heuristics (\foldan), no clear trend emerges once deviations are introduced. With heuristics (\foldah), the relationship appears polynomial or exponential. The differing coefficients between the model types indicate their influence on \foldan and \foldah.

\mysubsubsection{Impact of Deviations and Heuristics} 
The top row in Fig.~\ref{fig:sp_v_heuristic} shows that for \foldan, the number of visited states increases as the position of deviations moves toward the end of the trace.  In comparison, the bottom row shows that for \foldah, visited states remain consistent regardless of the positioning of deviations, indicating the heuristic helps control state-space explosion.

However, heuristics do not eliminate backtracking, and real-world performance may vary. Although the results show a linear relationship between synchronous product size and visited states, there seems to be a polynomial or exponential relation between visited states and elapsed time (Fig.~\ref{fig:v_t_heuristic}). 
Additionally, while heuristics mitigate the impact of deviations on time, it comes with a cost: for non-deviation cases, heuristics increase maximum computation time from 2.98s to 70.44s.

\mysubsubsection{Deviation Placement And Backtracking}
Backtracking is an inherent aspect of directed search, where the algorithm prioritizes lower-cost extensions, even if they are not immediately adjacent to the end state. The nonlinear relationship as observed in Fig.~\ref{fig:v_t_heuristic} confirms this challenge. 
Unlike traditional methods, unfolding alignments rely on a branching process, which requires the checking of multiple condition sets per transition to determine whether the transition can be a valid extension. \ch{As the branching process increases in complexity, the computational overhead of checking the condition sets rises exponentially, which likely explains the observed nonlinear relationship between visited states and elapsed time.}


\mysubsubsection{Impact Of Process Constructs} 
Fig.~\ref{fig:v_t_heuristic} indicates that EN-type models (nested choice) exhibit the fastest increase in elapsed time. Furthermore, when heuristics are applied, C and CN-type models display similar performance, suggesting that concurrency is managed more effectively than choice constructs. The introduction of choice constructs increases the number of branching configurations, thereby raising computational costs and degrading overall performance.

The findings highlight the significant impact of deviation placement on computational performance when no heuristics are used. While the number of visited states scales linearly with synchronous product size, elapsed time follows a polynomial or exponential relation when heuristics are used. Process models with choice and nested choice constructs impose the highest computational burden. 

\subsection{Experiment 2 - Comprehensive Performance Comparison}

This experiment evaluates \ch{\foldah using 19 log-model pairs.} For the 7 real-world event logs from various sectors such as healthcare and banking, for each of these logs, we discovered a normative process model using two algorithms: Inductive Miner (IM) with noise thresholds of 0.2 (except 0.5 for SP, as lower values did not result in a deviation in every trace) and Split Miner (SM) with default settings, (excluding the BPIC19 log, where SM did not return a model). Additionally, we also include 6 artificial event logs (ITL), which were used as a benchmark for large-scale conformance checking~\cite{munoz2013conformance}. For the ITL logs, we use the predefined reference models (275-429 transitions)~\cite{munoz2013conformance}. An overview is listed in Table~\ref{tab:meanmed}. 


Due to the dataset size, 1000 traces were randomly sampled from BPIC12, BPIC13, BPIC17, BPIC19, HB, and TR. Alignments were computed with a 100-second timeout per trace to ensure the feasibility of the experiment. 

\begin{table}[tb]
\caption{Elapsed times and queued and visited states of \foldah, \astar, and \dijk on each dataset and the corresponding process model. Note that the performance is calculated on the finished traces.}
\label{tab:meanmed}
\centering
%
\makebox[\linewidth]{ 

\resizebox{1.2\textwidth}{!}{%
\begin{tabular}{l|rrrr|rrr|rr|rr|rr|rrr|rrr}
              & \multicolumn{4}{c|}{Finished/total} & \multicolumn{3}{c|}{$\#$SPT} & \multicolumn{6}{c|}{Elapsed   time (sec.)}                                                  & \multicolumn{3}{c|}{Queued   states} & \multicolumn{3}{c}{Visited   states} \\
              \hline
              & \foldah  & Astar  & Dijkstra  & Total & \foldah & Astar & Dijkstra & \multicolumn{2}{c}{\foldah} & \multicolumn{2}{c}{Astar} & \multicolumn{2}{c}{Dijkstra} & \multicolumn{1}{|c}{\foldah }    & Astar     & Dijkstra     & \foldah      & Astar     & Dijkstra     \\
              \hline
Log-model pair       &       &        &           &       &      &       &          & Median      & Mean       & Median       & Mean       & Median        & Mean         & Mean     & Mean      & Mean         & Mean      & Mean      & Mean         \\
BPIC12         & 799   & 1000   & 1000      & 1000  & 92   & 106   & 106      & 0.102       & 2.014      & 0.015        & 1.440      & 0.633         & 1.795        & 274      & 14016     & 86252        & 144       & 3573      & 12867        \\
BPIC12\_SPLIT & 827   & 1000   & 1000      & 1000  & 64   & 77    & 77       & 0.013       & 0.954      & 0.004        & 0.037      & 0.007         & 0.011        & 156      & 206       & 680          & 87        & 115       & 375          \\
BPIC13        & 650   & 1000   & 1000      & 1000  & 31   & 31    & 31       & 0.013       & 0.039      & 0.002        & 0.003      & 0.002         & 0.002        & 57       & 88        & 161          & 31        & 32        & 62           \\
BPIC13\_SPLIT & 998   & 1000   & 1000      & 1000  & 25   & 25    & 25       & 0.008       & 0.139      & 0.002        & 0.004      & 0.001         & 0.002        & 84       & 44        & 86           & 66        & 28        & 55           \\
BPIC17        & 679   & 999    & 999       & 1000  & 135  & 151   & 151      & 3.135       & 15.607     & 0.941        & 2.349      & 0.237         & 0.524        & 2043     & 7784      & 21622        & 1346      & 2987      & 6946         \\
BPIC17\_SPLIT & 972   & 1000   & 1000      & 1000  & 115  & 117   & 117      & 0.198       & 1.230      & 0.010        & 0.021      & 0.020         & 0.026        & 233      & 162       & 1486         & 134       & 67        & 909          \\
BPIC19        & 986   & 1000   & 1000      & 1000  & 102  & 103   & 103      & 0.512       & 0.777      & 0.093        & 0.265      & 0.732         & 1.839        & 585      & 2577      & 70655        & 220       & 476       & 9476         \\
HB            & 1000  & 1000   & 1000      & 1000  & 67   & 67    & 67       & 0.049       & 0.055      & 0.004        & 0.006      & 0.003         & 0.003        & 74       & 96        & 229          & 33        & 29        & 78           \\
HB\_SPLIT     & 1000  & 1000   & 1000      & 1000  & 41   & 41    & 41       & 0.015       & 0.077      & 0.002        & 0.002      & 0.001         & 0.002        & 32       & 28        & 54           & 15        & 11        & 26           \\
SP            & 1048  & 1050   & 1050      & 1050  & 55   & 56    & 56       & 0.040       & 0.204      & 0.007        & 0.012      & 0.096         & 0.128        & 75       & 177       & 5425         & 39        & 32        & 1465         \\
SP\_SPLIT     & 1022  & 1050   & 1050      & 1050  & 53   & 55    & 55       & 0.041       & 0.895      & 0.007        & 0.150      & 0.110         & 0.141        & 122      & 492       & 6919         & 65        & 159       & 1974         \\
TR            & 1000  & 1000   & 1000      & 1000  & 35   & 35    & 35       & 0.015       & 0.012      & 0.002        & 0.002      & 0.057         & 0.041        & 27       & 55        & 2412         & 14        & 12        & 390          \\
TR\_SPLIT     & 1000  & 1000   & 1000      & 1000  & 26   & 26    & 26       & 0.007       & 0.006      & 0.001        & 0.001      & 0.001         & 0.001        & 17       & 17        & 36           & 8         & 6         & 18           \\
ITL prAm6     & 1200  & 1200   & 1200      & 1200  & 428  & 428   & 428      & 1.522       & 2.264      & 0.036        & 0.136      & 1.567         & 4.790        & 146      & 187       & 173193       & 62        & 58        & 34059        \\
ITL prBm6     & 1200  & 1200   & 1200      & 1200  & 402  & 402   & 402      & 1.947       & 1.926      & 0.034        & 0.053      & 0.031         & 0.047        & 137      & 169       & 169          & 43        & 41        & 41           \\
ITL prCm6     & 328   & 500    & 500       & 500   & 395  & 405   & 405      & 12.960      & 18.694     & 2.049        & 4.833      & 0.961         & 0.995        & 991      & 3794      & 63766        & 743       & 1614      & 18855        \\
ITL prDm6     & 732   & 732    & 0         & 1200  & 919  & 919   &  -        & 45.656      & 46.503     & 0.957        & 1.478      &     -          &      -        & 768      & 5216      &      -        & 264       & 257       &   -           \\
ITL prEm6     & 1200  & 1200   & 0         & 1200  & 475  & 475   &  -        & 5.456       & 5.493      & 0.115        & 0.148      &    -          &      -        & 323      & 1000      &       -       & 105       & 104       &   -           \\
ITL prFm6     & 686   & 686    & 0         & 1200  & 783  & 783   &   -       & 30.517      & 30.641     & 0.395        & 0.500      &     -          &      -        & 740      & 2894      &     -         & 247       & 245       &    -         
\end{tabular}%
}
}
\end{table}

\mysubsubsection{Results}
Table \ref{tab:meanmed} lists that for \foldah, the median of the elapsed time is much lower than the mean, with the exception of ITL prBm6, TR, and TR\_SPLIT. This difference indicates a possible long-tailed elapsed time distribution. We highlight ITL prCm6 and BPIC17, which very clearly exhibit this effect.
%
%
\foldah achieved sub-second mean computational times on 10 out of 13 real-life datasets (i.e., SP, SP\_SPLIT, BPIC19, HB, HB\_SPLIT, BPIC13, BPIC13\_SPLIT, TR, TR\_SPLIT, and BPIC12\_SPLIT). For the remaining datasets, mean times ranged from 1.23 to 46.50 seconds, with the highest elapsed times observed in BPIC12, BPIC17, BPIC17\_SPLIT, and the artificially generated ITL datasets.

When comparing performance, \astar consistently outperformed \foldah in elapsed time, requiring 71.45\% to 99.11\% less time across all completed datasets. However, \astar queued more states in 15 out of 19 datasets, ranging from 22.73\% to 578.80\% more while visiting fewer states in 16 out of 19 cases (0.80\% to 59.89\% fewer). 
%
Dijkstra also showed faster execution on \toch{13 out of 19} log-model pairs, reducing the elapsed time by 38.64\% to 99.27\%. However, it queued significantly more states across all datasets, with increases ranging from 1.89\% to 118,318.48\%, and visited more states in 14 out of 19 cases. 

Looking beyond aggregates, \foldah generally struggles with larger models and traces. It also suffers extreme outliers, often reaching time-outs on datasets that pose no problem to either \astar or \dijk. This is especially noticeable in the ITL prCm6 and BPIC17 datasets, where \foldah only complete 328/500 and 697/1000 traces, compared to the 500/500 and 1000/1000 for \astar and \dijk. 

Even in datasets without time-outs, \foldah exhibited significantly higher maximum elapsed times. For example, on SP and SP\_SPLIT, the mean times were 0.204 and 0.895 seconds, but the maximums reached 93.447 and 99.987 seconds, respectively. In contrast, the maximums required by \astar were 1.495 and 53.633 seconds, while \dijk needed only 2.820 and 4.056 seconds. 

\begin{figure}
   \includegraphics[width=\textwidth]{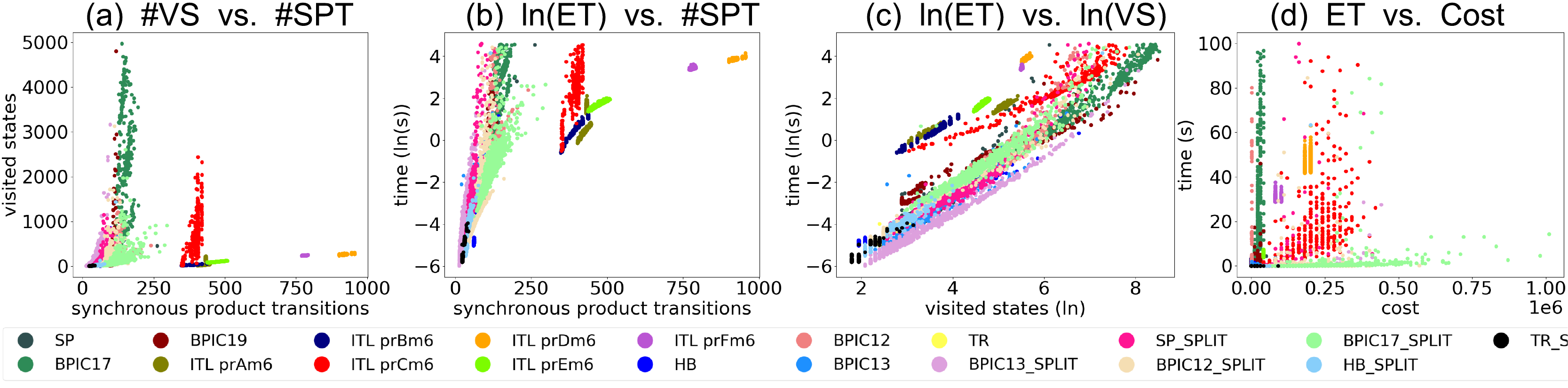}
    \caption{\ch{Correlation analysis of visited states, transitions, and elapsed time.}}
    \label{fig:st2_overview}
\end{figure}

Fig.~\ref{fig:st2_overview}(a) shows the effect of the number of transitions in the synchronous product on the visited states across datasets, demonstrating significant variations in visited states for a given synchronous product size and dataset. This variation can also be seen for the logarithm of elapsed time in Fig.~\ref{fig:st2_overview}(b). BPIC17 and ITL prCm6 stand out in particular.

Fig.~\ref{fig:st2_overview}(c) depicts the relationship between visited states and elapsed time, with the log-log plot suggesting a polynomial correlation. Similarly, Fig.~\ref{fig:st2_overview}(d) shows a linear relationship between synchronous product size and the logarithm of elapsed time, indicating an exponential dependence between them. Note that in both cases the precise relationship appears to be specific to a given model, indicating that, on their own, these metrics do not describe the performance characteristics generally. Additionally, this confirms that the previously assumed linear relationship between visited states and synchronous product size does not hold in more complex datasets.
Fig.~\ref{fig:st2_overview}(d) examines cost in relation to elapsed time but reveals no clear patterns due to high variability. 

Overall, these findings indicate that the outliers cannot be attributed entirely to a single metric. This tracks with the findings in Experiment 1, which suggest that the interplay between model complexity and the efficacy of the directed search drives performance degradation through the additional costs incurred by backtracking when compared to traditional approaches.


\mysubsubsection{Regression analysis} We applied linear regression to predict the elapsed time of \foldah, examining whether time complexity can be estimated using pre-computation metrics and if additional post-computation metrics improve the prediction. We use the $ln(ET)$ as our dependent variable. We selected the independent variables using correlation analysis and recursive feature elimination. The first regression, using only pre-computation metrics, yields a statistically significant model with an adjusted $R^2$ of 0.770. \ch{The estimated regression coefficients for each independent variable are}: $\#SPT (2.891)$, choice factor (0.715), and concurrency factor (-0.161). We define the choice factor as the average number of outgoing arcs per place in the model, and the concurrency factor as the average outgoing arcs per transition in the model. The second regression, incorporating post-computation metrics, achieves a stronger fit with an adjusted $R^2$ of 0.970. Its coefficients are: $\#SPT$ (1.288), visited states (ln) (-0.399), and queued states (ln) (2.271).
These findings, particularly those of the second regression, provide additional context for the performance characteristics of \foldah. Firstly, it confirms the impact model size has on performance. Secondly, it explains the impact of the directed search component via the relation between the coefficient of visited states and queued states. In essence, the negative coefficient for visited states results in lower elapsed times when a greater portion of the states that were queued, are also visited. It thereby proxies the efficiency of the directed search component of the algorithm.

\section{Conclusion}\label{sec:conclusion}
This paper introduced a new approach for computing optimal partial-order alignments using Petri net unfoldings. We designed and implemented an algorithm that integrates Petri net unfoldings with a directed search, evaluating its performance across different process constructs and comparing it to traditional alignment methods.

Our findings indicate that unfolding alignments are computationally viable but sensitive to model structure and deviation placement, with choice constructs and late deviations leading to performance degradation. We mitigated this with a heuristic-guided search, improving efficiency. While our approach underperforms compared to AStar and Dijkstra in terms of execution time, it does effectively reduce the number of queued states. Moreover, this approach retains dependencies between moves during the unfolding, which allows for designing more advanced cost functions in the future, improving diagnostic capabilities beyond traditional sequential alignment approaches. It can also be easily adapted to incremental and parallel alignment computing. Future research may also focus on optimizing performance by designing heuristics tailored towards branching processes. 

\mysubsubsection{Acknowledgement}
We thank Artem Polyvyanyy, Dirk Fahland, and Thomas Chatain for their invaluable discussion in 2017 on the previous idea that inspired this idea.

\bibliographystyle{splncs04}
\bibliography{bib}
%





\newpage
\appendix
\section{Additional Examples}
\label{app:example}
\subsection{Partial-order traces}
\begin{figure}[h]
    \centering
    \includegraphics[width=0.8\linewidth]{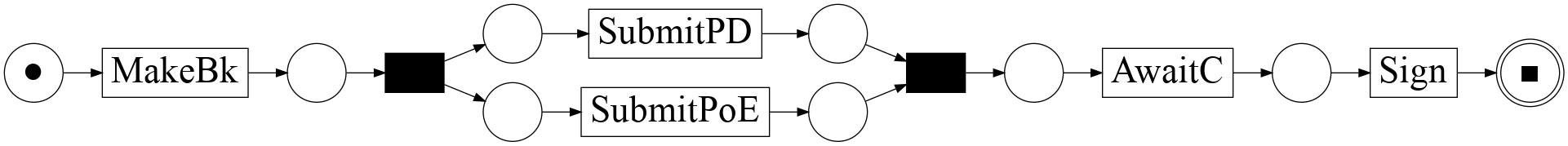}
    \caption{The partial order trace, converted to an event net.}
    \label{fig:po_en}
    \centering
    \includegraphics[width=0.8\linewidth]{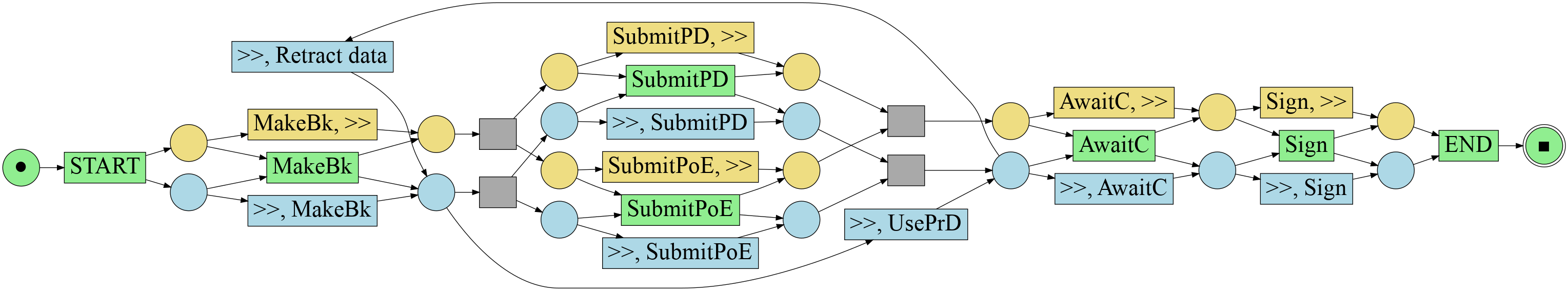}
    \caption{The synchronous product.}
    \label{fig:po_sp}
    \centering
    \includegraphics[width=0.8\linewidth]{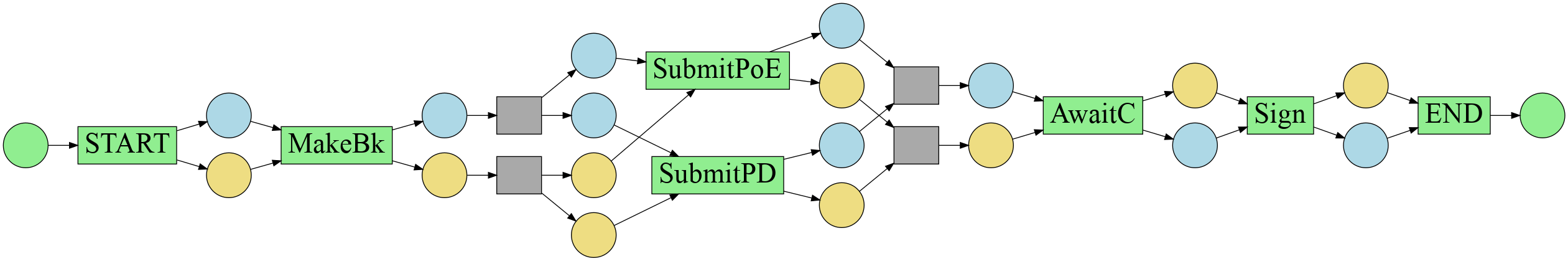}
    \caption{The unfolding alignment.}
    \label{fig:po_sp}
\end{figure}

\subsection{Unbounded, easy sound net}

\begin{figure}[H]
    \centering
    \includegraphics[width=0.5\linewidth]{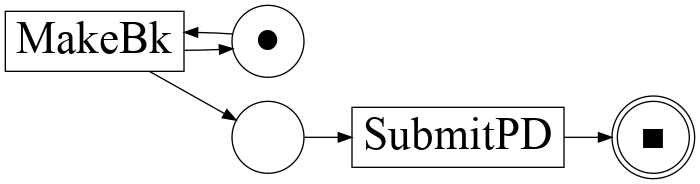}
    \caption{An unbounded, easy sound net, meaning the final marking is reachable, where the final marking is a token in the initial place and a token in the final place.}
    \label{fig:unboundednet}
\end{figure}

\begin{figure}[H]
    \centering
    \includegraphics[width=0.8\linewidth]{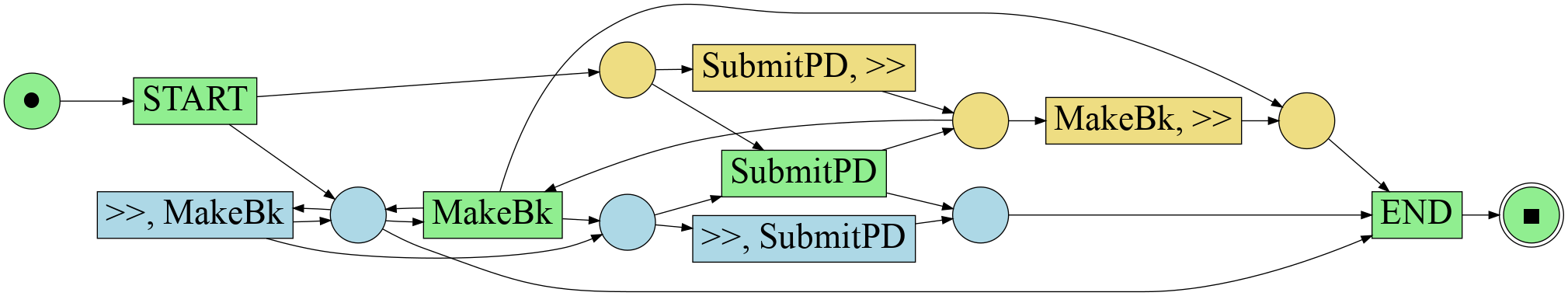}
    \caption{The synchronous product between the easy sound net and the trace $\sigma = \langle \acsubmitpay, \acmakebook\rangle$. Note that we added a dummy start and a dummy end transition to ease the checking of termination conditions.}
    \label{fig:unbounded_sp}
\end{figure}

\begin{figure}[H]
    \centering
    \includegraphics[width=0.8\linewidth]{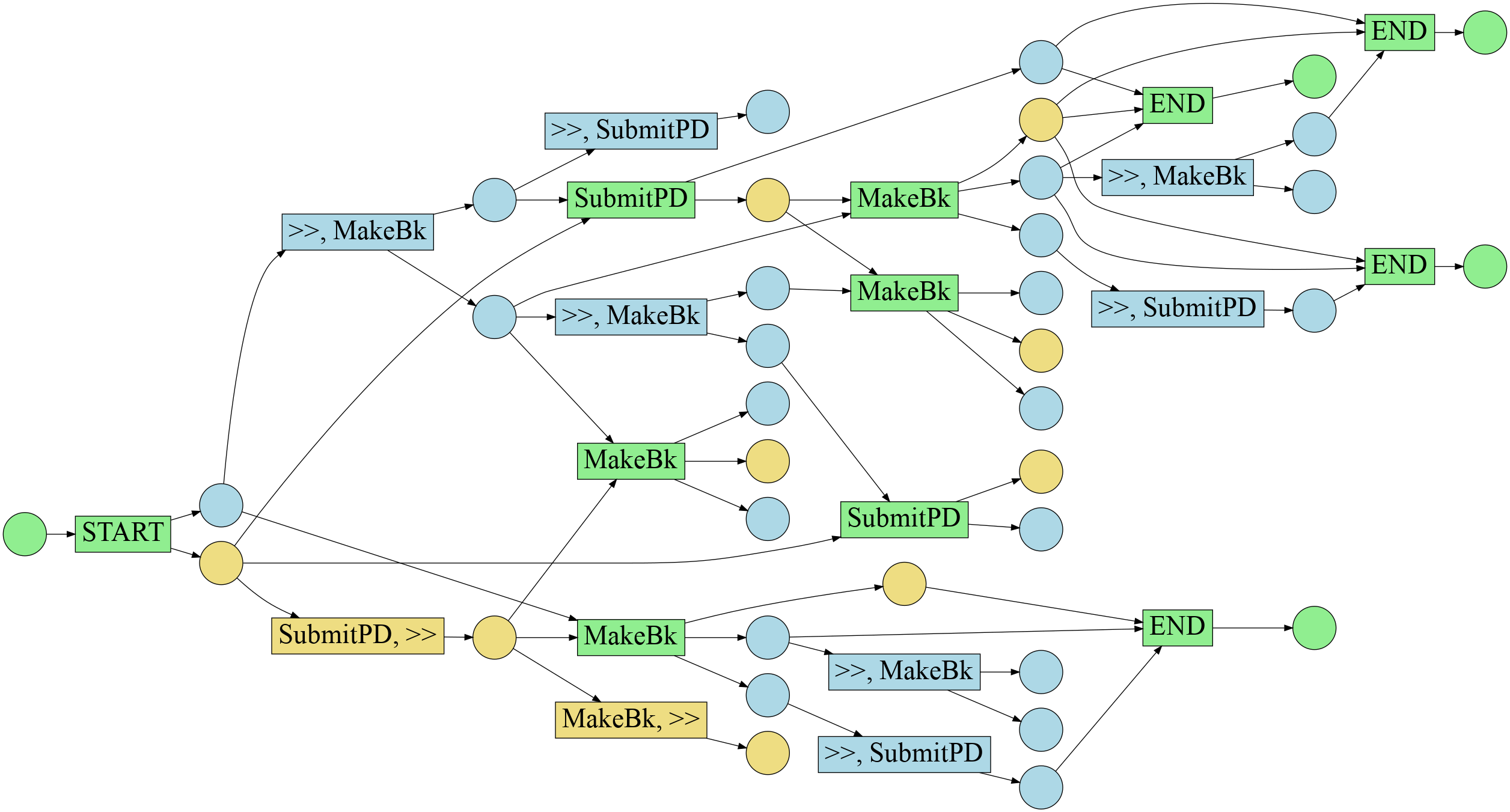}
    \caption{The directed unfolding, trying to find the target marking.}
    \label{fig:unbound_unf}
\end{figure}

\begin{figure}[H]
    \centering
    \includegraphics[width=0.8\linewidth]{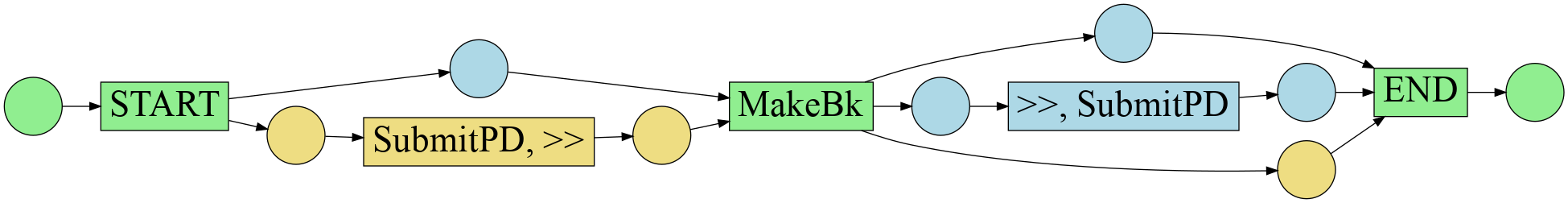}
    \caption{The final unfolding alignment net.}
    \label{fig:unbounded_alignment}
\end{figure}

\section{Notations}
\label{sec:notation}

A \textbf{multiset} $b$ over $A$ is a function $b : A \rightarrow \mathbb{N}$, which returns the count of element $a$ in multiset $b$. We use $\mathcal{B}(A)$ for the set of all multisets over a finite domain $A$. 
For example, $b_1 = []$ (empty), $b_2 = [x, x, y]$, $b_3 = [x,y,z]$, and $b_4 = [x^3, y^2, z]$ are multisets over $A = \{x, y, z\}$, where order does not matter. 
Standard set operations extend to multisets, e.g., union $(b_2 \uplus b_3 = b_4)$, difference $(b_4 \setminus b_2 = b_3)$, and cardinality $(|b_4| = 6)$. A multiset $b$ is a subset of $b'$ $(b \leq b')$ if $b(a) \leq b'(a)$ for all $a \in A$, with strict inclusion $(b < b')$ requiring inequality for at least one element. 
%
%

A \textbf{sequence} $\sigma = \langle a_1, a_2, ..., a_n \rangle$ over $X$ has length $|\sigma|$ and supports concatenation, e.g., $\langle x, x, y \rangle \cdot \langle x, y, z \rangle = \langle x, x, y, x, y, z \rangle$. Sequences can be converted to multisets using $[a \in \sigma]$, e.g., $[a \in \langle x, x, y, x, y, z \rangle ] = [x^3, y^2, z]$. 

\begin{definition}[Petri Net] A \term{Petri Net} $N = (P, T, F)$ is a 3-tuple in which $P$ is a set of places, $T$ is a set of transitions, and $F \subseteq (P \times T) \cup (T \times P)$ defines the flow relations. A \term{marking} $m$ of $N$ is a multiset of $P$, i.e., $m$ assigns each place $p \in P$ a number $m(p)$ of tokens and presents a state of the net. 

A \term{marked Petri} net $N = (P, T, F, i,f)$ is a Petri net with an \term{initial marking} $i$ and a \emph{final marking} $f$.  
The \term{preset} $\preset{x}$ of a node $x$ is the set $\{ y \mid (y, x) \in F\}$. Similarly, the \term{postset} $\postset{x}$ of a node $x$ is the set $\{ y \mid (x, y) \in F \}$.

A transition $t$ of $N$ is \term{enabled} at a marking $m$ of $N$, iff for all $p \in \preset{p}$, $m(p) > 0$. If $t$ is enabled at $m$, then $t$ may \term{fire} or occur, and its occurrence leads to a successor marking $m'$ with $m'(p) = m(p) - \preset{t} + \postset{t}$. We use $m \xrightarrow{t} m'$ to denote this relation. A sequence $\sigma = \langle t_1, t_2, \cdots, t_n \rangle$ of transitions is an \term{occurrence sequence} of $N$ if there exist markings $i, m_1, m_2, ..., m_n$ such that $i \xrightarrow{t_1} m_1 \xrightarrow{t_2} m_2 \cdots \xrightarrow{t_n} m_n$, also denoted by $i \xrightarrow{\sigma} m_n$. A marking $m$ is a \term{reachable} marking if there exists an occurrence sequence $\sigma$ such that $i \xrightarrow{\sigma} m$. A marking $M$ of a net is \term{n-safe} if $m(p) \leq n$ for every place $p \in P$. A net is n-safe if all its reachable markings are n-safe. In the following, we assume all nets to be 1-safe nets. 
\end{definition}

\begin{definition}[Causal $<$, conflict $\#$, and concurrency $||$] Given a net, we define the causal, conflict, and concurrency relations between nodes of a net as follows:
\begin{itemize}
    \item Two nodes $x$ and $y$ are in \emph{causal relation}, denoted by $x < y$, if the net contains a path with at least one arc leading from $x$ to $y$.
    \item $x$ and $y$ are in\emph{conflict relation}, or just in conflict, denoted by $x \# y$, if the net contains two paths $st_1, \cdots, x_1$ and $st_2, \cdots, x_2$ starting at the same place $s$, and such that $t_1 = t_2$. In words, $x_1$ and $x_2$ are in conflict if the net contains two paths leading to $x_1$ and $x_2$ which start at the same place and immediately diverge (although later on they can converge again).
    \item $x$ and $y$ are in \emph{concurrency relation}, denoted by $x || y$, if neither $x < y$ nor $y < x$ nor $x \# y$.
\end{itemize}
\end{definition}

\end{document}